\documentclass[10pt,journal,compsoc]{IEEEtran}
\ifCLASSOPTIONcompsoc
  \usepackage[nocompress]{cite}
\else
  \usepackage{cite}
\fi
\ifCLASSINFOpdf
\else
\fi
\usepackage{amsmath}
\usepackage{amsfonts}
\usepackage{graphicx}
\usepackage{multirow}
\usepackage{booktabs}
\usepackage[table]{xcolor}
\usepackage{listings}
\usepackage{algorithm}
\usepackage{algorithmic}

\usepackage{ragged2e}
\usepackage[colorlinks,linkcolor=red]{hyperref}

\begin{document}
%
\title{USTEP: Spatio-Temporal Predictive Learning under A Unified View}
%
%
%
%

\author{Cheng~Tan$^*$,
        Jue~Wang$^*$,
        Zhangyang~Gao$^*$,
        Siyuan Li,
        and~Stan~Z.~Li,~\IEEEmembership{Fellow,~IEEE}
\IEEEcompsocitemizethanks{
\IEEEcompsocthanksitem * Equal contribution. \protect\\
\vspace{-2mm}
\IEEEcompsocthanksitem Cheng Tan, Jue Wang, Zhangyang Gao and Siyuan Li are with Zhejiang University, Hangzhou, China, and also with the AI Lab, Research Center for Industries of the Future, Westlake University. Email: \{tancheng, wangjue, gaozhangyang\}@westlake.edu.cn.
\IEEEcompsocthanksitem Stan Z. Li is with the AI Lab, Research Center for Industries of the Future, Westlake University. Email: Stan.ZQ.Li@westlake.edu.cn. \protect\\
}
}

%
%

\markboth{Journal of \LaTeX\ Class Files,~Vol.~14, No.~8, July~2022}%
{Shell \MakeLowercase{\textit{et al.}}: Bare Demo of IEEEtran.cls for Computer Society Journals}
%



\IEEEtitleabstractindextext{%
\begin{abstract}
Spatio-temporal predictive learning plays a crucial role in self-supervised learning, with wide-ranging applications across a diverse range of fields. Previous approaches for temporal modeling fall into two categories: recurrent-based and recurrent-free methods. The former, while meticulously processing frames one by one, neglect short-term spatio-temporal information redundancies, leading to inefficiencies. The latter naively stack frames sequentially, overlooking the inherent temporal dependencies. In this paper, we re-examine the two dominant temporal modeling approaches within the realm of spatio-temporal predictive learning, offering a unified perspective. Building upon this analysis, we introduce USTEP (Unified Spatio-TEmporal Predictive learning), an innovative framework that reconciles the recurrent-based and recurrent-free methods by integrating both micro-temporal and macro-temporal scales. Extensive experiments on a wide range of spatio-temporal predictive learning demonstrate that USTEP achieves significant improvements over existing temporal modeling approaches, thereby establishing it as a robust solution for a wide range of spatio-temporal applications.
\end{abstract}

}

\maketitle

\IEEEdisplaynontitleabstractindextext

%
\IEEEpeerreviewmaketitle

\IEEEraisesectionheading{\section{Introduction}\label{sec:introduction}}

%
%
%
%

 

\IEEEPARstart{I}{n} an era where data is continually streaming in, there is an increasing demand to not only understand the present but to also predict the future. By leveraging historical video data, spatio-temporal predictive learning strives to forecast subsequent sequences in an unsupervised manner~\cite{finn2016unsupervised,ddpae,locatello2019challenging,greff2019multi,mathieu2019disentangling,khemakhem2020variational,castrejon2019improved,oord2016wavenet,shen2020timeseries}. With real-world applications extending from forecasting weather patterns~\cite{convlstm,earthformer,rasp2020weatherbench,ning2023mimo,seo2023implicit} to predicting traffic flows~\cite{fang2019gstnet,wang2019memory} and simulating physical interactions~\cite{lerer2016learning,finn2016unsupervised}, the ramifications of advancements in this promising domain are profound.

\begin{figure}[h]
\centering
\vspace{-2mm}
\includegraphics[width=0.98\linewidth]{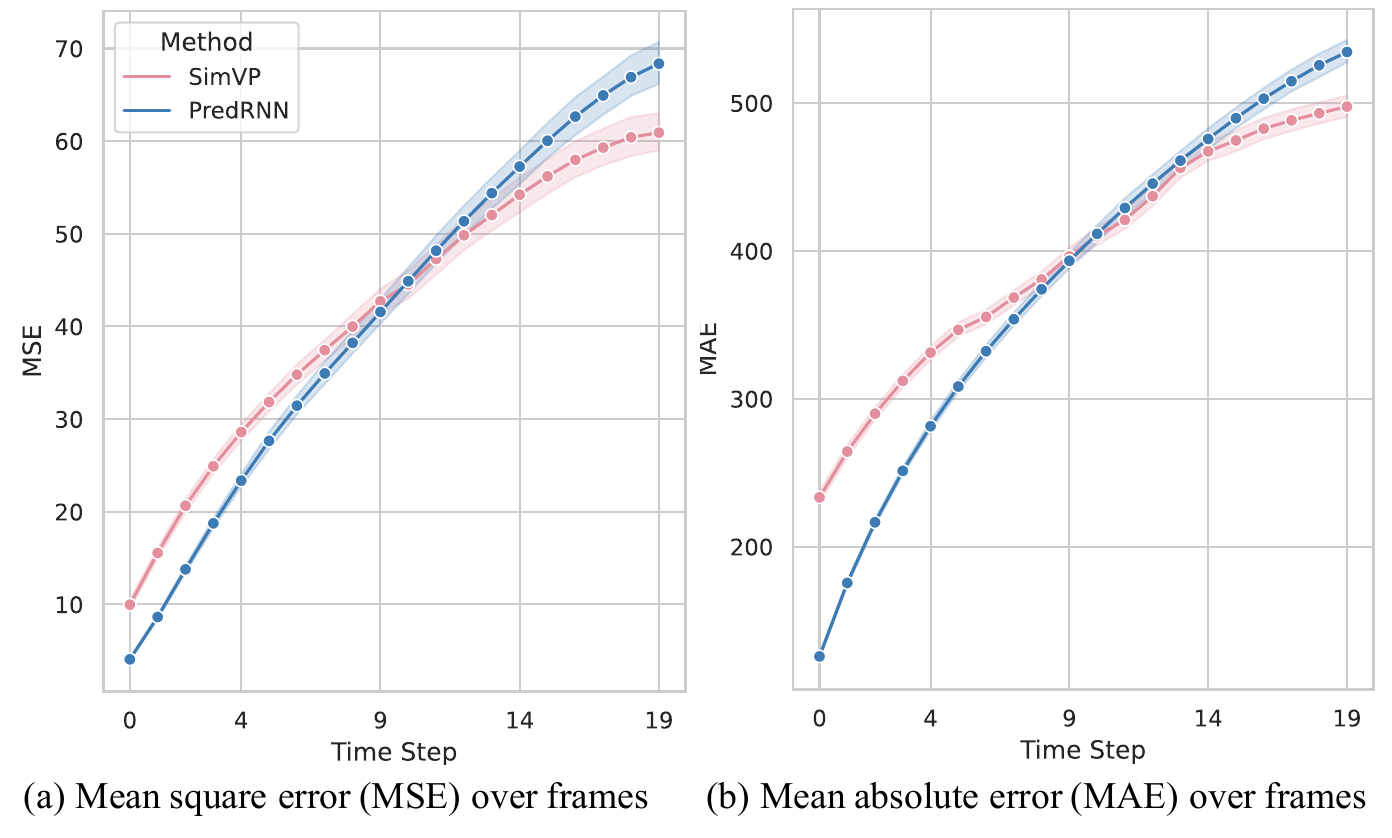}
\vspace{-2mm}
\caption{Frame-by-frame MSE/MAE comparison between the representative recurrent-based method PredRNN and the recurrent-free method SimVP on the extended frame task using the KTH dataset. The plot illustrates the differences in performance across individual frames.}
\label{fig:claim}
\end{figure}

The path to achieving accurate spatio-temporal predictions has been fraught with challenges. Traditional approaches have typically oscillated between two primary temporal modeling methodologies: recurrent-based and recurrent-free methods. The recurrent-based methods~\cite{convlstm,predrnn,prednet,predrnn++,predrnnv2,mim,e3dlstm,jin2020exploring,babaeizadeh2021fitvid} meticulously process frames one by one, ensuring that temporal relationships across each timestep are captured. Yet, they often grapple with inefficiencies arising from the redundant short-term spatio-temporal information and challenges in preserving global information from preceding time steps. Conversely, the recurrent-free methods~\cite{tan2023temporal,simvp,simvpv2,tan2023openstl}, while alleviating the inefficiencies of their recurrent counterparts, fall short in capturing the inherent temporal dependencies. By stacking frames in a naive manner, these models may overlook the intricate dance of cause and effect played out over time, risking missing the subtle interplay. As shown in Figure~\ref{fig:claim}, a frame-by-frame MSE/MAE comparison on the KTH dataset illustrates the strengths and weaknesses of these two approaches. Although the overall average performance metrics for the recurrent-based method PredRNN~\cite{predrnn} and the recurrent-free method SimVP~\cite{simvp} are nearly identical (MSE 41.07 vs. 41.11, MAE 380.6 vs. 397.1), a more granular, frame-by-frame analysis reveals critical insights. PredRNN outperforms SimVP in the initial frames, highlighting its effectiveness in capturing short-term dependencies. However, in the latter frames, SimVP shows superior performance, underscoring its ability to handle long-term dependencies. The frame-by-frame comparison thus emphasizes the complementary nature of these two approaches. PredRNN's strength in modeling short-term dependencies and SimVP's capability to manage long-term dependencies highlight the inherent trade-offs in choosing one method over the other. This observation underscores the necessity for a unified framework that integrates the advantages of both methodologies, addressing their shortcomings.

In this work, we revisit the foundational principles of temporal modeling in spatio-temporal predictive learning, dissecting the merits and demerits of the prevailing approaches. We introduce the concept of a temporal segment, defined as a subsequence encompassing a series of continuous frames. To refine our understanding further, we formally identify and delineate two temporal scales: the micro-temporal scale, which focuses on immediate, sequential dependencies, and the macro-temporal scale, which encapsulates long-range global patterns. Recurrent-based methods primarily concentrate on micro-temporal scales, adeptly capturing instantaneous interactions but often lacking in long-term insight. Conversely, recurrent-free methods excel in considering macro-temporal scales, effectively capturing broader temporal patterns, but their neglect of immediate temporal dependencies results in a loss of richness in the predicted sequences. This discrepancy between the two paradigms highlights a significant gap in current methodologies. The key challenge is reconciling the redundant per-frame processing of recurrent-based methods with the naive stacking of recurrent-free methods into a unified spatio-temporal learning framework that can effectively model both micro- and macro-temporal scales.

\begin{figure}[h]
\centering
\includegraphics[width=0.98\linewidth]{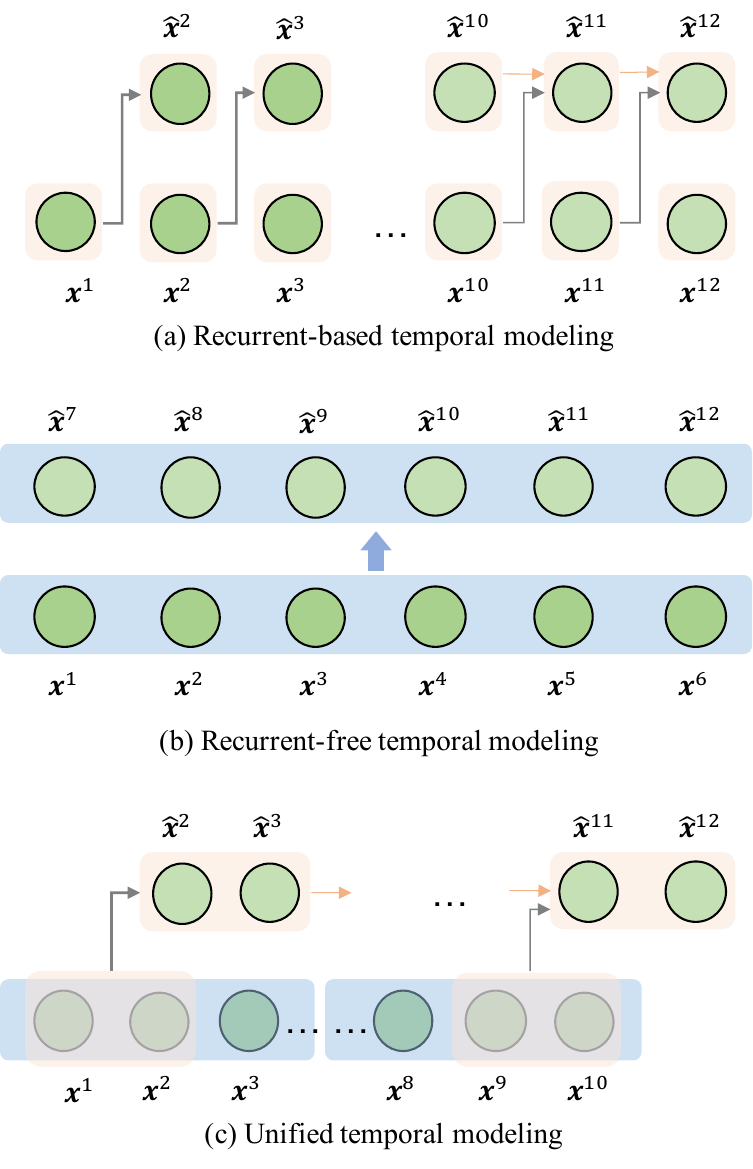}
\caption{Temporal modeling comparison between recurrent-based, recurrent-free and our unified temporal modeling.}
\label{fig:temporal_comparison}
\end{figure}

To bridge this gap, we introduce USTEP (Unified Spatio-TEmporal Predictive learning), a novel framework that takes into account both micro- and macro-temporal scales. By doing so, USTEP achieves a balanced trade-off between predictive performance and computational efficiency. The architecture of USTEP is designed to integrate seamlessly the strengths of both recurrent-based and recurrent-free methods, while also introducing new mechanisms that enhance the model's ability to generalize across various spatio-temporal scales. As illustrated in Fig.~\ref{fig:temporal_comparison}, recurrent-based methods focus solely on the micro-temporal scale, processing one frame at a time; recurrent-free methods consider only the macro-temporal scale, treating the entire sequence as a single unit. USTEP defines micro-temporal segments as several consecutive frames and considers the entire input sequence as a macro-temporal segment, maintaining long-range dependencies as context. This dual-scale approach allows USTEP to learn and update both short-term and long-term temporal relationships dynamically. We conduct a diverse range of spatio-temporal predictive tasks and the experimental results demonstrate its superior performance, not just in terms of accuracy but also in computational efficiency. We find that USTEP achieves state-of-the-art performance with moderate computational resources. Such efficiency and effectiveness establish USTEP as an exceptionally powerful solution, particularly suited to addressing the multifaceted challenges of real-world applications.

A preliminary version of this work was published in~\cite{tan2023temporal}, focusing on advancing temporal attention, and on the exploration of open-source software for spatiotemporal learning~\cite{tan2023openstl} (\url{github.com/chengtan9907/OpenSTL}). This current journal paper extends them in the following aspects: 
\vspace{-1mm}
\begin{enumerate}
  \item Leveraging insights gained from the comprehensive spatiotemporal benchmark in OpenSTL~\cite{tan2023openstl}, this paper provides an in-depth review of prevailing approaches to spatiotemporal predictive learning. A novel perspective is introduced by categorizing these methods according to their operational scales—specifically, micro-temporal and macro-temporal scales. This classification enables a nuanced analysis of their limitations and strengths, fostering a more coherent understanding of the landscape of spatiotemporal predictive learning.
  \item Addressing the dichotomy between recurrent-based and recurrent-free learning paradigms, we propose USTEP—a novel, integrated framework designed to harmonize these methodologies by incorporating both micro-temporal and macro-temporal considerations. Using TAU~\cite{tan2023temporal} as the basic unit, USTEP represents a significant methodological advancement, offering a versatile and robust platform for spatiotemporal predictive analysis that transcends the limitations of existing approaches.
  \item The efficacy and efficiency of USTEP are rigorously validated across a broad spectrum of spatiotemporal predictive learning tasks. Our experimental investigations encompass tasks with equal frame, extended frame, and reduced frame requirements, demonstrating USTEP's superior performance and adaptability across varying temporal dynamics and spatial configurations. 
\end{enumerate}

\section{Related Work}

\subsection{Recurrent-based Method}
Recurrent-based models have made significant strides in spatio-temporal predictive learning. Drawn inspiration from recurrent neural networks~\cite{hochreiter1997long}, VideoModeling~\cite{marc2014video} incorporates language modeling techniques and employs quantization of patches into a dictionary for recurrent units. ConvLSTM~\cite{convlstm} leverages convolutional neural networks to model the LSTM architecture. PredNet~\cite{prednet} persistently predicts future video frames using deep recurrent convolutional neural networks with bottom-up and top-down connections. PredRNN~\cite{predrnn} proposes a Spatio-temporal LSTM (ST-LSTM) unit that extracts and memorizes spatial and temporal representations simultaneously, and its following work PredRNN++~\cite{predrnn++} further introduces gradient highway unit and Casual LSTM to capture temporal dependencies adaptively. E3D-LSTM~\cite{e3dlstm} designs eidetic memory transition in recurrent convolutional units. PredRNN-v2~\cite{predrnnv2} has expanded upon PredRNN by incorporating a memory decoupling loss and a curriculum learning technique. However, these models struggle with capturing long-term dependencies. Moreover, they tend to be computationally intensive, especially when scaled to high-dimensional data, limiting their practical applicability.

\subsection{Recurrent-free Method}

Instead of employing computationally intensive recurrent methods for spatio-temporal predictive learning, alternative approaches such as PredCNN~\cite{predcnn} and TrajectoryCNN~\cite{liu2020trajectorycnn} utilize convolutional neural networks for temporal modeling. SimVP~\cite{simvp,simvpv2} represents a seminal work that incorporates blocks of Inception modules within a UNet architecture. In parallel, the introduction of the TAU~\cite{tan2023temporal} represents another leap forward. This innovation underscores the critical role of attention mechanisms in achieving more nuanced and efficient temporal understanding in neural network models. Despite these advances, it is imperative to acknowledge the inherent limitations that accompany these models, particularly in their ability to capture the intricacies of fine-grained temporal dependencies. Additionally, a notable constraint arises from the models' design philosophy concerning output length. Predominantly, these architectures are configured to produce outputs that mirror the length of their inputs, a design choice that, while offering a measure of structural symmetry, introduces rigidity and inherently limits the models' flexibility in decoding.

\subsection{Efficient Recurrent Neural Network}

In sequence modeling, RWKV~\cite{peng2023rwkv} and RetNet~\cite{sun2023retentive} revisited the potential of RNNs and propose RNN architectures that can achieve performance comparable to Transformers~\cite{vaswani2017attention}. Mega~\cite{ma2022mega} proposes a chunk-wise recurrent design, using the moving average equipped gated attention mechanics to capture long-range dependencies in sequential data across various modalities. This design is particularly adept at capturing long-range dependencies across a diverse array of data modalities, highlighting the evolving capabilities of recurrent models in managing complex sequential information. While these prior works demonstrate that well-designed recurrent architectures can be both effective and efficient, USTEP goes a step further by synergizing recurrent and recurrent-free paradigms. This hybrid approach allows USTEP to capture both micro- and macro-temporal scales, offering a nuanced and robust framework for spatio-temporal predictive learning.

\vspace{-4mm}
\section{Background}

We formally define the spatio-temporal predictive learning problem, inspired by existing works~\cite{simvp,tan2023openstl}. Consider an observed sequence of frames $\mathcal{X}^{t, T} = \{\boldsymbol{x}^i\}_{t-T+1}^t$ at a specific time $t$, comprising the past $T$ frames. Our objective is to forecast the subsequent $T'$ frames, denoted as $\mathcal{Y}^{t+1, T'} = \{\boldsymbol{x}^{i}\}_{t+1}^{t+T'}$. Each frame $\boldsymbol{x}_i$ is generally an image in $\mathbb{R}^{C \times H \times W}$, with $C$ being the number of channels, $H$ the height, and $W$ the width. In the tensorial representation, the observed and predicted sequences are represented as $\mathcal{X}^{t, T} \in \mathbb{R}^{T \times C \times H \times W}$ and $\mathcal{Y}^{t+1, T'} \in \mathbb{R}^{T' \times C \times H \times W}$.

Given a model with learnable parameters $\Theta$, we seek to find a mapping $\mathcal{F}_\Theta: \mathcal{X}^{t, T} \mapsto \mathcal{Y}^{t+1, T'}$. This mapping is realized through a neural network model that captures both spatial and temporal dependencies within the data. The model is trained to minimize a loss function $\mathcal{L}$ that quantifies the discrepancy between the predicted and ground-truth future frames, formulated as:
\begin{equation}
  \min_{\Theta} \mathcal{L}(\mathcal{F}_\Theta(\mathcal{X}^{t, T}), \mathcal{Y}^{t+1, T'}),
\end{equation}
where $\mathcal{L}$ is chosen to evaluate the quality of the predictions in both spatial and temporal dimensions. With the formal problem definition in place, the key challenges in spatio-temporal predictive learning lie in the effective modeling of temporal dependencies, which can be summarized as:
\begin{itemize}
  \item \textit{Short-term redundancies removal:} Recurrent-based methods process frames one by one, ensuring that temporal relationships across each timestep are captured. However, it often leads to inefficiencies due to redundant processing of short-term spatio-temporal information. These redundancies can cause unnecessary computational overhead and hinder the model's ability to focus on significant temporal changes.
  \item \textit{Long-term context preservation:} Maintaining global information from preceding time steps is crucial for accurate long-term predictions. Recurrent-based methods struggle with this as they tend to lose global context. Conversely, recurrent-free methods, which stack frames in a sequence, are more efficient but fail to capture the intricate temporal dependencies necessary for understanding long-term context. This can result in predictions that miss the nuanced interplay of events over time, leading to less accurate forecasts.
\end{itemize}

To address these challenges, we define a temporal segment as a basic unit comprising a series of continuous frames. We introduce two distinct temporal scales to differentiate between short-term dynamics and long-term context: the micro-temporal scale and the macro-temporal scale. The micro-temporal scale focuses on individual basic temporal segments, capturing fine-grained and immediate interactions. In contrast, the macro-temporal scale encompasses the context with the entire input sequence length, providing a broader context and capturing long-term dependencies. The detailed definitions of them are as follows.

\noindent{\textbf{Definition 3.1}} (temporal segment). \textit{A temporal segment is defined as a contiguous subsequence of frames extracted from a given spatio-temporal sequence for the purpose of efficient temporal modeling. Formally, let $U_j = \{\boldsymbol{x}^i\}_{t_j}^{t_j+\Delta t-1}$, where $t_j$ is the starting time of the segment and $\Delta t$ is the length of the segment measured in time units or number of frames.}

By focusing on temporal segments, we aim to capture essential temporal features while mitigating spatio-temporal redundancies that often occur in short-term sequences.

\noindent{\textbf{Definition 3.2}} (micro-temporal scale). \textit{The micro-temporal scale refers to the granularity at which a spatio-temporal sequence is partitioned into non-overlapping, contiguous temporal segments for the purpose of efficient and localized temporal modeling. Formally, a sequence $\{\boldsymbol{x}^i\}_{t-T+1}^{t+T'}$ is divided into $N$ micro-temporal segments $\mathcal{U} = \{U_1, U_2, ..., U_N\}$.}

The micro-temporal scale divides the spatio-temporal sequence into non-overlapping temporal segments, emphasizing the independent nature of each segment. This approach captures fine-grained features within each segment while avoiding overlap, thereby ensuring that the detailed information within short-term sequences is accurately preserved and efficiently processed.

\noindent{\textbf{Definition 3.3}} (macro-temporal scale). \textit{The macro-temporal scale refers to the granularity at which a spatio-temporal sequence is divided into large temporal segments, each encompassing the length of the entire input sequence. Formally, a sequence $\{\boldsymbol{x}^i\}_{t-T+1}^{t+T'}$ is divided into $M$ macro-temporal segments $\mathcal{V} = \{V_1, V_2, ..., V_M\}$, where each segment $V_j = \{U_{j1}, U_{j2}, ..., U_{jk}\}$ consists of $k$ micro-temporal segments, and $k\Delta t = \Delta T$ with $\Delta T = T \gg \Delta t$.}

The macro-temporal scale aims to capture long-term context by encompassing segments that span the entire length of the input spatio-temporal sequence. The initial macro-temporal segment is equivalent to the recurrent-free approach, providing a broad overview of the input sequence. Subsequent macro-temporal segments advance with a step size equal to that of the micro-temporal segments by updating the long-term context, ensuring a comprehensive and continuous understanding of long-term dependencies across the entire sequence.

\begin{figure}[ht]
\centering
\includegraphics[width=0.98\linewidth]{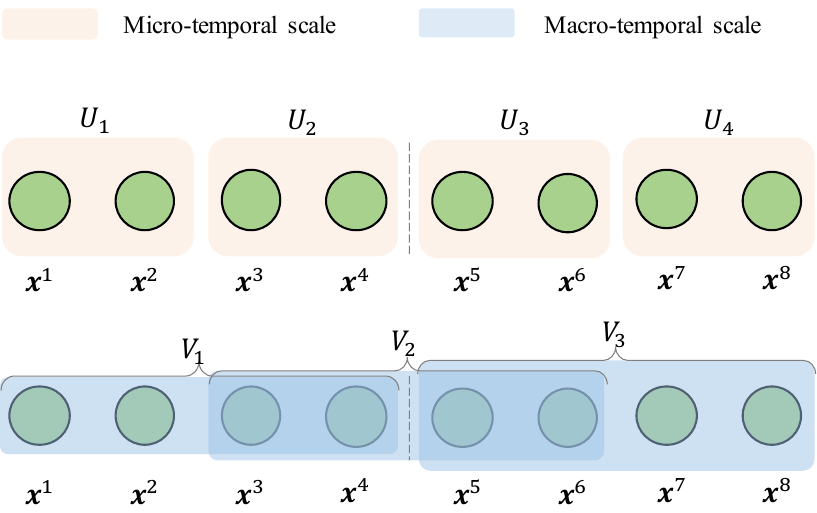}
\caption{The illustration of micro- and macro-temporal scales. Here we take a $4 \rightarrow 4$ frames prediction as an example. Each green circle represents an individual frame. Micro-temporal scales (\textit{in red}) divide the sequence into non-overlapping temporal segments, containing a few consecutive frames. Macro-temporal scale (\textit{in blue}) further divides the sequence into large temporal segments. The number of $\mathcal{V}$ is $|\mathcal{U}|-1$.}
\label{fig:temporal_scale}
\end{figure}

\subsection{Recurrent-based Temporal Modeling}

In recurrent-based temporal modeling~\cite{convlstm,predrnn,phydnet}, there exists a predominant focus on the intricacies of micro-temporal scales, to the exclusion of broader, macro-temporal dynamics. Formally, each micro-temporal segment $U_i$ consists solely of a single frame $\{\boldsymbol{x}^i\}$ with $\Delta t = 1$. The modeling approach can be expressed as follows:
\begin{equation}
  \widehat{U}_{i+1} = \begin{cases}
    \mathcal{F}_\Theta(U_i, H_{i-1}), & \text{if }  t-T+1 \leq i \leq t , \\
    \mathcal{F}_\Theta(\widehat{U}_i, H_{i-1}), & \text{otherwise},
  \end{cases}
\end{equation}
where $H_{i-1}$ is the hidden state from the preceding frame. The model operates in two distinct phases: 
\begin{itemize}
  \item \textit{Reconstruction Phase:} For historical frames $(t-T+1 \leq i \leq t)$, the model learns to reconstruct the next frame $\widehat{U}_{i+1}$ based on the current ground-truth frame $U_i$ and the hidden state $H_{i-1}$ from the preceding frame.
  \item \textit{Prediction Phase:} For future frames $i > t$, the model uses the hidden state $H_{i-1}$ and the last predicted frame $\widehat{U}_i$ to predict the next frame $\widehat{U}_{i+1}$.
\end{itemize}

In both phases, the efficacy of the model for either reconstructing or predicting frames is contingent upon the effective and straightforward learning of the hidden state $H_{i-1}$ from the preceding frame. 

\subsection{Recurrent-free Temporal Modeling}

In recurrent-free temporal modeling~\cite{simvp,tan2023temporal}, the focus shifts entirely to the macro-temporal scale, bypassing any micro-temporal segments. Specifically, each macro-temporal segment $V$ is defined as a sequence of $T$ consecutive frames, with $\Delta T = T$. The recurrent-free temporal modeling approach can be mathematically expressed as follows:
\begin{equation}
  \widehat{V}_2 = \mathcal{F}_\Theta(V_1),
\end{equation}
where $V_1 = \{\boldsymbol{x}^i \}_{t-T+1}^{t}$ is the historical frames, and $V_2 = \{\boldsymbol{x}^i \}_{t+1}^{t+T'}$ is the ground-truth future frames, $\widehat{V}_2$ is the predicted future frames by the model $\mathcal{F}_\Theta$. The model operates in a single phase, where the model learns to predict the future frames $\widehat{V}_2$ based on the historical frames $V_1$. It is worth noting that here the output frames have the same length as the input frames.

By working with macro-temporal segments, recurrent-free temporal modeling exhibits computational advantages, as it can process multiple frames in parallel. It excels in capturing global patterns over the entire temporal window by taking the macro-temporal segment into account. However, it falters in handling intricate temporal dependencies, primarily because it lacks micro-temporal granularity. The fixed size of macro-temporal segments is inflexible and limits the practical applicability.

\subsection{Summary} 

In summary, we have dissected the foundational principles underlying temporal modeling in both recurrent-based and recurrent-free methodologies. These approaches are distinguished by their primary operational focus on different temporal scales—namely, the micro-temporal and macro-temporal scales. While recurrent-based temporal modeling focuses on micro-temporal scales, recurrent-free temporal modeling shifts focus to the macro-temporal scale, processing sequences of frames as a whole. 

\begin{figure*}[h]
\centering
\includegraphics[width=0.98\textwidth]{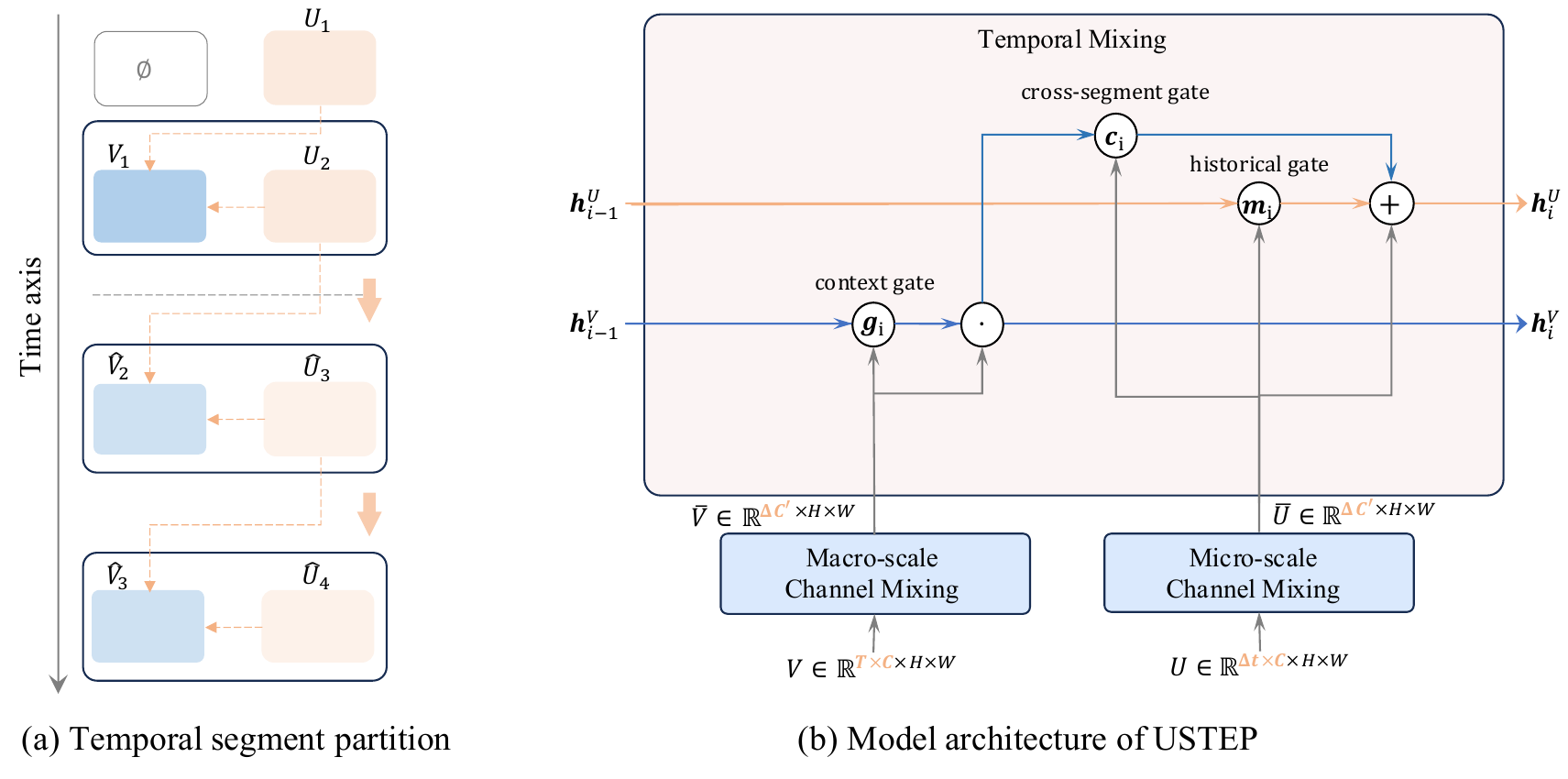}
\caption{Illustration of USTEP’s unified spatio-temporal predictive learning framework. (a) Temporal segment partition: USTEP constructs macro-temporal segments using both the previous and the current micro-temporal segments. This approach allows the model to eliminate short-term redundancies while preserving long-term context. (b) Detailed architecture of USTEP: The framework consists of two specialized recurrent-free modules, $F^U_{\theta_1}$ and $F^V_{\theta_2}$, which handle channel mixing for micro- and macro-temporal scales, respectively. Hidden states from both scales are integrated through a gating mechanism and cross-segment-level temporal modeling, ensuring comprehensive spatio-temporal predictive learning.}
\label{fig:ustep_temporal}
\end{figure*}

\section{USTEP: Unified Spatio-Temporal Predictive Learning}

To overcome the limitations inherent in both recurrent-based and recurrent-free modeling, we propose USTEP, a unified framework designed to harmoniously integrate micro- and macro-temporal scales, aiming to maximize the effectiveness of spatio-temporal predictive learning.

\subsection{Temporal Scale Sets}

The initial stage in USTEP involves dividing the input frame sequence into two separate sets, corresponding to the micro-temporal and macro-temporal scales, which are denoted as $\mathcal{U}$ and $\mathcal{V}$, respectively, as illustrated in Fig.~\ref{fig:temporal_scale}. In the micro-temporal scale set $\mathcal{U}$, each temporal segment $U_i$ is constructed to contain a few consecutive frames, facilitating the capture of fine-grained spatio-temporal information. The length of each $U_i$ is determined by $\Delta t$, which is chosen to balance the trade-off between temporal granularity and computational efficiency. For the macro-temporal scale set $\mathcal{V}$, we employ a sliding window approach to construct larger temporal segments. Each macro-temporal segment $V_i$ contains multiple non-overlapping segments $U_i$. The sliding window moves in steps of size $\Delta t$, ensuring that the macro-temporal segments are constructed in a manner consistent with the micro-temporal scale. As shown in Fig.~\ref{fig:ustep_temporal}(a), the macro-temporal segments are constructed using both the previous and the current micro-temporal segments, enabling the model to eliminate short-term redundancies while preserving long-term context. This hierarchical approach ensures that the model can effectively capture both immediate and long-term temporal dependencies.

\subsection{Single Segment-Level Temporal Modeling}

Upon segmenting the input frame sequence into the micro- and macro-temporal sets, $\mathcal{U}$ and $\mathcal{V}$, the next step is to perform temporal modeling at the single-segment level. This division engages in temporal modeling at an individual segment level, with a concerted effort to achieve a harmonious representation across both temporal dimensions. The core ambition of this phase is to derive hidden states from each temporal segment that are not only compatible across the micro- and macro-temporal scales but are also cohesively aligned within the same feature space that maintains uniform dimensionality in a recurrent-free manner.

To facilitate this objective, temporal segments $U_i$ and $V_i$, representative of both the micro and macro scales, respectively, are transposed into a shared feature domain, articulated with dimensions $C' \times H \times W$. This transposition is executed through the deployment of two specialized recurrent-free modules, denoted as $F^U_{\theta_1}$ and $F^V_{\theta_2}$. Each module is parameterized with a set of learnable parameters $\theta_1, \theta_2$. These modules transform the original segments to corresponding hidden states, denoted as $\overline{U}_i, \overline{V}_i \in \mathbb{R}^{C' \times H \times W}$, according to the following equations:
\begin{equation}
  \overline{U}_i = F^U_{\theta}(U_i), \quad \overline{V}_i = F^V_{\theta}(V_i),
\end{equation}
By congruently aligning both micro- and macro-temporal segments within the same feature dimension $C'$, we establish a unified representation space where the hidden states from both temporal scales can be directly integrated. Such a unified feature space enables us to leverage the complementary strengths of micro- and macro-temporal modeling, thus enhancing the overall performance. Analogous to the concept of channel mixing in MetaFormer~\cite{yu2022metaformer}, single segment-level temporal modeling can be viewed as channel mixing at the micro- and macro-temporal scales individually. This approach allows for the independent processing of fine-grained and broad contextual information.

\vspace{-3mm}
\subsection{Cross Segment-Level Temporal Modeling}

Once the unified hidden states are obtained from both micro- and macro-temporal segments, the next challenge is to harmoniously integrate these at the cross-segment level. Our approach is to leverage the advantages of both micro- and macro-temporal scales, capturing fine-grained detail while maintaining a global perspective, to enhance the predictive capability of our model, USTEP.

The macro-temporal scale hidden states, $\overline{V}_i$, are processed using a gating mechanism as follows:
\begin{equation}
\centering
\begin{aligned}
  \boldsymbol{g}_{i} &= \sigma(W_{v} \ast \overline{V}_i + b_{v}),\\
  \boldsymbol{h}_{i}^V &= \overline{V}_i + \boldsymbol{g}_{i} \odot \boldsymbol{h}_{i-1}^V,
\end{aligned}
\end{equation}
where $\sigma(\cdot)$ denotes the Sigmoid activation function, $\ast$ and $\odot$ represent the convolution operator and the Hadamard product, respectively. The parameters $W_v$ and $b_v$ are the convolution weights and bias. The context gate, \(\boldsymbol{g}_{i}\), controls the flow of historical macro-temporal scale information and preserves long-term context by regulating the integration of new information with the previously accumulated macro-temporal scale hidden states.

Subsequently, the micro-temporal scale and the processed macro-temporal hidden states are then integrated:
\begin{equation}
\centering
\begin{aligned}
  \boldsymbol{m}_{i} &= \sigma(W_u \ast \overline{U}_i + b_{u}), \\
  \boldsymbol{c}_{i} &= \sigma(W_c \ast \overline{U}_i + b_{c}), \\
  \boldsymbol{h}_i^U &= \overline{U}_i + \boldsymbol{m}_i \odot \boldsymbol{h}_{i-1}^U + \boldsymbol{c}_i \odot \boldsymbol{h}_{i}^V,
\end{aligned}
\end{equation}
where $W_u, W_c, b_u,$ and $b_c$ are the convolution weights and biases. The historical gate $\boldsymbol{m}_i$ controls the integration of the historical micro-temporal information, ensuring that the model effectively utilizes past micro-temporal hidden states. The cross-segment gate $\boldsymbol{c}_i$ manages the incorporation of the macro-temporal hidden state, ensuring that long-term context is integrated with the immediate temporal features. 
This dual-gate mechanism ensures that the model leverages both short-term and long-term dependencies, 
enhancing its predictive capabilities and overall performance.

The detailed schematic illustration of the temporal modeling learning process within USTEP is depicted in Fig.~\ref{fig:ustep_temporal}(b). For every single segment, regardless of whether it belongs to the micro-temporal or macro-temporal scale, a recurrent-free approach is employed to swiftly capture coarse temporal dependencies. In contrast, when considering the relationships across segments, a recurrent-based approach is sequentially applied to discern finer temporal dependencies. Notably, during this stage, temporal dependencies from different scales are harmoniously fused, ensuring a comprehensive temporal understanding. This hierarchical and integrative approach allows USTEP to achieve a delicate balance between capturing immediate temporal nuances and understanding broader temporal patterns. 
  
In the training phase, our proposed USTEP method adopts a partitioning approach to divide the input sequences into $x$ and $y$, which aligns with the training strategy typically employed by recurrent-based methods. However, USTEP offers a flexible approach compared to traditional methods, which often use a fixed time step of $\Delta t = 1$. The flexibility in USTEP allows for different time step sizes, accommodating various temporal resolutions and capturing temporal dependencies at different granularities. During the inference phase, the model engages in iterative operations for a predetermined number of steps. These steps encompass both those required for observing sequences $x$ and the actual steps required for generating the predictions.

The algorithmic structure for the training process of our proposed USTEP is presented in Algorithm~\ref{alg:train}, providing a comprehensive roadmap for its implementation. Notably, the partition function supports to handle with videos with flexible lengths by adding up a pad. 

\begin{algorithm}[H]
\caption{Pseudocode of training}
\label{alg:train}
\definecolor{codeblue}{rgb}{0.25,0.5,0.5}
\definecolor{codekw}{rgb}{0.85, 0.18, 0.50}
\lstset{
  backgroundcolor=\color{white},
  basicstyle=\fontsize{7.5pt}{7.5pt}\ttfamily\selectfont,
  columns=fullflexible,
  breaklines=true,
  captionpos=b,
  commentstyle=\fontsize{7.5pt}{7.5pt}\color{codeblue},
  keywordstyle=\fontsize{7.5pt}{7.5pt}\color{codekw},
}
\begin{lstlisting}[language=python]
def train(data, delta_t, delta_T):
      x = data[:, :-delta_t]
      y = data[:, delta_t:]
      # partition: micro and macro sets
      u, v = partition(x, delta_t, delta_T)

      # single segment-level: recurrent-free
      u, v = f_u(u), f_v(v)
      # cross segment-level: recurrent-based
      pred = []
      h_v, h_u = 0, 0
      for i in range(len(u)-1):
          h_v = macro_func(v[i], h_v)
          h_u = micro_func(u[i+1], h_u, h_v)
          pred.append(h_u)
      
      loss = loss_func(pred, y)
      return loss
\end{lstlisting}
\end{algorithm}

During the inference phase, USTEP engages in a series of iterative operations spanning a pre-defined number of steps, which include both the observation of sequence $x$ and the execution of steps necessary for generating forward-looking predictions. This iterative approach ensures a dynamic and responsive prediction process, tailored to the temporal characteristics of the input data.

The essence of this inference process is captured within the pseudocode provided in Algorithm~\ref{alg:infer} below. This algorithm is not just a sequence of steps but a manifestation of the USTEP framework's core capabilities, illustrating its approach to integrating and operationalizing the micro and macro temporal scales in a seamless and efficient manner.

\begin{algorithm}[H]
  \caption{Pseudocode of inference}
  \label{alg:infer}
  \definecolor{codeblue}{rgb}{0.25,0.5,0.5}
  \definecolor{codekw}{rgb}{0.85, 0.18, 0.50}
  \lstset{
    backgroundcolor=\color{white},
    basicstyle=\fontsize{7.5pt}{7.5pt}\ttfamily\selectfont,
    columns=fullflexible,
    breaklines=true,
    captionpos=b,
    commentstyle=\fontsize{7.5pt}{7.5pt}\color{codeblue},
    keywordstyle=\fontsize{7.5pt}{7.5pt}\color{codekw},
  }
  \begin{lstlisting}[language=python]
  def inference(x, delta_t, delta_T, n_step):
        # x: B x T x C x H x W
        # partition: micro and macro sets
        u, v = partition(x, delta_t, delta_T)
  
        # single segment-level: recurrent-free
        u, v = f_u(u), f_v(v)
        # cross segment-level: recurrent-based
        pred = []
        h_v, h_u = 0, 0
        for i in range(n_step-1):
            h_v = macro_func(v[i], h_v)
            h_u = micro_func(u[i+1], h_u, h_v)
            # practical prediction
            if i >= len(u):
              pred.append(h_u)
  
        return pred
\end{lstlisting}
\end{algorithm}

\section{Experiments}
\label{sec:exp}

We evaluate the efficacy of USTEP across three prevalent types of spatiotemporal prediction tasks:
\begin{itemize}
  \item \textit{Equal Frame Task:} The number of output frames matches that of the input frames. This task type inherently favors recurrent-free temporal modeling approaches due to its structured nature.
  \item \textit{Extended Frame Task:} The count of output frames substantially surpasses that of the input frames. This type of task is generally more compatible with recurrent-based temporal modeling approaches, allowing for more flexible, frame-by-frame predictions.
  \item \textit{Reduced Frame Task:} Diverging from the former, this task necessitates fewer output frames than the input frames. By mitigating the impact of cumulative errors, this task directly evaluates the model's capability in learning historical frames.
\end{itemize}

\noindent{\textbf{Datasets}} To rigorously assess the performance and applicability, we undertake a comprehensive quantitative evaluation across a diverse range of datasets, meticulously curated to encompass both synthetic and real-world scenarios:
\begin{itemize}
  \item \textbf{Moving MNIST}~\cite{srivastava2015unsupervised} is a synthetic dataset consisting of two digits moving within the 64 $\times$ 64 grid and bouncing off the boundary. It is a standard benchmark in spatiotemporal predictive learning.
  \item \textbf{Human 3.6M}~\cite{ionescu2013human3}  is a 3D human motion capture dataset for fitness, close human interactions, and self-contact. This dataset contains 3.6 million human poses and corresponding images, 11 professional actors (6 male, 5 female), and 17 scenarios (discussion, smoking, taking photos, talking on the phone, etc.). 
  \item \textbf{Weather Benchmark}~\cite{rasp2020weatherbench} This dataset contains various types of climatic data from 1979 to 2018. The raw data is regrind to low resolutions, we here choose $5.625^{\circ}$ ($32 \times 64$ grid points) resolution for our data. Since the complete data is very large and includes massive climatic attributes like geopotential, temperature, and other variables, we specifically chose the global temperature prediction task in evaluation.
  \item \textbf{Caltech Pedestrian} is a driving dataset focusing on detecting pedestrians. It consists of approximately 10 hours of $640\times 480$ videos taken from vehicles driving through regular traffic in an urban environment. We follow the protocol of PredNet~\cite{prednet} and CrevNet~\cite{crevnet} for pre-processing, training, and evaluation.
  \item \textbf{SEVIR}~\cite{veillette2020sevir} is a comprehensive storm event imagery dataset comprising over 10,000 weather events. Each event is represented by image sequences covering a spatial extent of 384 km $\times$ 384 km and spanning a temporal duration of 4 hours.
  \item \textbf{UCF Sports}~\cite{rodriguez2008action}  is a dataset featuring a collection of human actions performed in various sports scenarios. These actions are typically recorded from footage broadcast on television channels.
  \item \textbf{KTH}~\cite{schuldt2004recognizing} contains 25 individuals performing six types of actions. Following~\cite{villegas2017decomposing,e3dlstm}, we use persons 1-16 for training and 17-25 for testing. Models are trained to predict the next 20 frames from the previous 10 observations.
\end{itemize}
We summarize the statistics of the above datasets in Table~\ref{tab:dataset}, including the number of training samples $N_{train}$ and the number of testing samples $N_{test}$.

\begin{table}[h]
\centering
\setlength{\tabcolsep}{2.4mm}{
\caption{The statistics of datasets. The training or testing set has $N_{train}$ or $N_{test}$ samples, composed by $T$ or $T'$ images with the shape $(C, H, W)$.}
\label{tab:dataset}
\begin{tabular}{ccccccc}
\toprule
& $N_{train}$ & $N_{test}$ & $(C,H, W)$ & $T$ & $T'$ \\
\midrule
MMNIST    &  10,000 &  10,000 & (1, 64, 64)   & 10  & 10 \\
Human 3.6M &  73,404 &  8,582  & (3, 256, 256)   & 4   & 4 \\
WeatherBench & 2,167 & 706 & (1, 32, 64) & 12& 12 \\
Kitti\&Caltech & 3,160 & 3,095 & (3, 128, 160) & 10 & 1 \\
SEVIR    &  35,718  &  12,159  & (1, 384, 384) & 13  & 12 \\
UCF Sports & 6,288  & 752  & (3, 480, 720) & 4  & 1 \\
KTH       &  4,940  &  3,030  & (1, 128, 128) & 10  & 20/40 \\
\bottomrule
\end{tabular}}
\end{table}

\noindent\textbf{Baselines} We choose the following baselines for comparison: (i) Recurrent-based methods including ConvLSTM~\cite{convlstm}, SV2P~\cite{babaeizadeh2018stochastic}, SAVP~\cite{lee2018stochastic}, PredRNN~\cite{predrnn}, PredRNN++~\cite{predrnn++}, MIM~\cite{mim}, E3D-LSTM~\cite{e3dlstm}, and PredRNNv2~\cite{predrnnv2}; (ii) Recurrent-free methods including SimVP~\cite{simvp}, TAU~\cite{tan2023temporal}, Uniformer~\cite{li2022uniformer}, MLP-Mixer~\cite{tolstikhin2021mlp}, ConvNeXt~\cite{liu2022convnet} and IAM4VP~\cite{seo2023implicit}. For USTEP, we utilize TAU, SimVP, and Uniformer as the single segment-level temporal modeling modules $F^U_{\theta_1}$ and $F^V_{\theta_2}$ to demonstrate the robustness and versatility of our method. Notably, in USTEP, these modules retain the same architecture as their original models but are configured with fewer channels to ensure a fair comparison under similar computational complexity. 

\noindent\textbf{Measurement} We employ Mean Squared Error (MSE), Mean Absolute Error (MAE), Structure Similarity Index Measure (SSIM), and Peak Signal to Noise Ratio (PSNR) to evaluate the quality of predictions. MSE and MAE estimate the absolute pixel-wise errors, SSIM measures the similarity of structural information within the spatial neighborhoods, and PSNR is an expression for the ratio between the maximum possible power of a signal and the power of distorted noise. LPIPS~\cite{zhang2018unreasonable} is a perceptual similarity metric that computes the distance between two images' feature representations in a pre-trained deep network. FVD score~\cite{unterthiner2019fvd} is also included to evaluate the performance on the KTH dataset.

\noindent\textbf{Implementation details} We implement the proposed method with the Pytorch framework and conduct experiments on a single NVIDIA-V100 GPU. The AdamW optimizer is utilized with a learning rate of 0.01 and a weight decay of 0.05. The learning rate is chosen from a set of values, $\{1e^{-2},5e^{-3},1e^{-3},5e^{-4},1e^{-4}\}$, and the best result for each experiment is reported. It is worth noting that the experimental results in TAU~\cite{tan2023temporal} replicate the experimental settings from the previous baselines, which are not fully rigorous. For consistency and comparability, we adhere to the same settings as OpenSTL~\cite{tan2023openstl}. 

\begin{table*}[!h]
  \centering
  \caption{The quantitative results of different approaches on the equal frame task. The units for Params and FLOPs are M and G. The results of the USTEP series that underperform the default are in \textcolor{blue}{blue} while outperforming those are in \textcolor{red}{red}.}
  \resizebox{0.96\textwidth}{0.47\textheight}{%
  \begin{tabular}{c|c|cccccc}
  \toprule
  \multicolumn{1}{c|}{Dataset} & \multicolumn{1}{c|}{Metric} & \multicolumn{5}{c}{Method} \\
  \midrule
  \multirow{21}{*}{\rotatebox{90}{Moving MNIST}}
  & & ConvLSTM & PredRNN & PredRNN++ & MIM & E3DLSTM \\
  & MSE & 29.80 & 23.97 & \underline{22.06} & 22.55 & 35.97 \\
  & MAE & 90.64 & 72.82 & \underline{69.58} & 69.97 & 78.28 \\
  & SSIM ($\times 10^{-2}$) & 92.88 & 94.62 & \underline{95.09} & 94.98 & 93.20 \\
  & PSNR & 22.10 & 23.28 & \underline{23.65} & 23.56 & 21.11 \\
  & Params & \textbf{15.0} & 23.8 & 38.6 & 38.0 & 51.0 \\
  & FLOPs & 56.8 & 116.0 & 171.7 & 179.2 & 298.9 \\ 
  \cline{2-7}
  \addlinespace[0.1cm]
  & & PredRNNv2 & MLP-Mixer & \cellcolor{gray!10} Uniformer & \cellcolor{gray!10} SimVP & \cellcolor{gray!10} TAU \\
  & MSE & 24.13 & 29.52 & \cellcolor{gray!10} 30.38 & \cellcolor{gray!10} 32.15 & \cellcolor{gray!10} 24.60 \\
  & MAE & 73.73 & 83.36 & \cellcolor{gray!10} 85.87 & \cellcolor{gray!10} 89.05 & \cellcolor{gray!10} 71.93 \\
  & SSIM ($\times 10^{-2}$) & 94.53 & 93.38 & \cellcolor{gray!10} 93.08 & \cellcolor{gray!10} 92.68 & \cellcolor{gray!10} 94.54\\
  & PSNR & 23.21 & 22.13 & \cellcolor{gray!10} 22.78 & \cellcolor{gray!10} 21.84 & \cellcolor{gray!10} 23.19 \\
  & Params & 23.9 & 44.8 & \cellcolor{gray!10} 46.8 & \cellcolor{gray!10} 58.0 & \cellcolor{gray!10} 44.7 \\
  & FLOPs  & 116.6 & 16.5 & \cellcolor{gray!10} 16.5 & \cellcolor{gray!10} 19.4 & \cellcolor{gray!10} 16.0 \\
  \cline{2-7}
  \addlinespace[0.1cm]
  &     & ConvNeXt & IAM4VP & \cellcolor[HTML]{E7ECE4}USTEP w/Uniformer & \cellcolor[HTML]{E7ECE4}USTEP w/SimVP & \cellcolor[HTML]{E7ECE4}USTEP w/TAU \\
  & MSE & 26.94 & 27.04 & \cellcolor[HTML]{E7ECE4} 28.51$_{(\textrm{\textcolor{red}{-1.87})}}$ & \cellcolor[HTML]{E7ECE4} 30.04$_{(\textrm{\textcolor{red}{-2.11})}}$ & \cellcolor[HTML]{E7ECE4}\textbf{21.84}$_{(\textrm{\textcolor{red}{-2.76})}}$ \\
  & MAE & 77.23 & 79.70 & \cellcolor[HTML]{E7ECE4} 82.43$_{(\textrm{\textcolor{red}{-3.44})}}$ & \cellcolor[HTML]{E7ECE4} 85.92$_{(\textrm{\textcolor{red}{-3.13})}}$ & \cellcolor[HTML]{E7ECE4}\textbf{63.21}$_{(\textrm{\textcolor{red}{-8.72})}}$  \\
  & SSIM ($\times 10^{-2}$) & 93.97 & 93.95 & \cellcolor[HTML]{E7ECE4} 94.32$_{(\textrm{\textcolor{red}{+1.24})}}$ & \cellcolor[HTML]{E7ECE4} 93.59$_{(\textrm{\textcolor{red}{+0.91})}}$ & \cellcolor[HTML]{E7ECE4}\textbf{95.38}$_{(\textrm{\textcolor{red}{+0.84})}}$   \\
  & PSNR   & 22.22 & 22.19 & \cellcolor[HTML]{E7ECE4} 23.28$_{(\textrm{\textcolor{red}{+0.50})}}$ & \cellcolor[HTML]{E7ECE4} 22.11$_{(\textrm{\textcolor{red}{+0.27})}}$ & \cellcolor[HTML]{E7ECE4}\textbf{24.06}$_{(\textrm{\textcolor{red}{+0.87})}}$  \\
  & Params & 37.3 & 35.4 & \cellcolor[HTML]{E7ECE4}\underline{17.8}$_{(\textrm{\textcolor{red}{-61.97\%})}}$ & \cellcolor[HTML]{E7ECE4}24.4$_{(\textrm{\textcolor{red}{-57.93\%})}}$ & \cellcolor[HTML]{E7ECE4}18.9$_{(\textrm{\textcolor{red}{-57.72\%})}}$ \\
  & FLOPs  & \textbf{14.1} & \underline{14.2} & \cellcolor[HTML]{E7ECE4}16.8$_{(\textrm{\textcolor{blue}{+1.82\%})}}$ & \cellcolor[HTML]{E7ECE4}21.9$_{(\textrm{\textcolor{blue}{+12.89\%})}}$ & \cellcolor[HTML]{E7ECE4}17.7$_{(\textrm{\textcolor{blue}{+10.63\%})}}$ \\
  \midrule
  \multirow{21}{*}{\rotatebox{90}{Human3.6M}} 
  & & ConvLSTM & PredRNN & PredRNN++ & MIM & E3DLSTM \\
  & MSE & 125.5 & 113.2 & 110.0 & 112.1 & 143.3 \\
  & MAE & 1566.7 & 1458.3 & 1452.2 & 1467.1 & 1442.5\\
  & SSIM ($\times 10^{-2}$) & 98.13 & 98.31 & 98.32 & 98.29 & 98.03 \\
  & PSNR & 33.40 & 33.94 & 34.02 & 33.97 & 32.52 \\
  & Params & 15.5 & 24.6 & 39.3 & 47.6 & 60.9 \\
  & FLOPs & 347.0 & 704.0 & 1033.0 & 1051.0 & 542.0 \\
  \cline{2-7}
  \addlinespace[0.1cm]
  & & PredRNNv2 & MLP-Mixer & \cellcolor{gray!10} Uniformer & \cellcolor{gray!10} SimVP & \cellcolor{gray!10} TAU \\
  & MSE & 114.9 & 116.3 & \cellcolor{gray!10} \underline{108.4} & \cellcolor{gray!10} 115.8 & \cellcolor{gray!10} 113.3 \\
  & MAE & 1484.7 & 1497.7 & \cellcolor{gray!10} 1441.0 & \cellcolor{gray!10} 1511.5 & \cellcolor{gray!10} \underline{1390.7} \\
  & SSIM ($\times 10^{-2}$) & 98.27 & 98.24 & \cellcolor{gray!10} 98.34 & \cellcolor{gray!10} 98.22 & \cellcolor{gray!10} 98.39 \\
  & PSNR & 33.84 & 33.76 & \cellcolor{gray!10}34.08 & \cellcolor{gray!10} 33.73 & \cellcolor{gray!10} 34.03 \\
  & Params & 24.6 & 27.7 & \cellcolor{gray!10}11.3  & \cellcolor{gray!10} 41.2 & \cellcolor{gray!10} 37.6 \\
  & FLOPs & 708.0 & 211.0 & \cellcolor{gray!10}74.6 & \cellcolor{gray!10} 197.0 & \cellcolor{gray!10} 182.0 \\
  \cline{2-7}
  \addlinespace[0.1cm]
  &     & ConvNeXt & IAM4VP & \cellcolor[HTML]{E7ECE4}USTEP w/Uniformer & \cellcolor[HTML]{E7ECE4}USTEP w/SimVP & \cellcolor[HTML]{E7ECE4}USTEP w/TAU \\
  & MSE & 113.4 & 114.7 & \cellcolor[HTML]{E7ECE4} \textbf{106.1}$_{(\textrm{\textcolor{red}{-2.3})}}$ & \cellcolor[HTML]{E7ECE4} 109.8$_{(\textrm{\textcolor{red}{-6.0})}}$ & \cellcolor[HTML]{E7ECE4}109.5$_{(\textrm{\textcolor{red}{-3.8})}}$ \\
  & MAE & 1469.7 & 1489.9 & \cellcolor[HTML]{E7ECE4} 1415.3$_{(\textrm{\textcolor{red}{-25.7})}}$ & \cellcolor[HTML]{E7ECE4} 1472.7$_{(\textrm{\textcolor{red}{-38.8})}}$ & \cellcolor[HTML]{E7ECE4}\textbf{1380.5}$_{(\textrm{\textcolor{red}{-10.2})}}$ \\
  & SSIM ($\times 10^{-2}$) & 98.28 & 98.26 & \cellcolor[HTML]{E7ECE4} \underline{98.42}$_{(\textrm{\textcolor{red}{+0.08})}}$ & \cellcolor[HTML]{E7ECE4} 98.38$_{(\textrm{\textcolor{red}{+0.16})}}$ & \cellcolor[HTML]{E7ECE4}\textbf{98.45}$_{(\textrm{\textcolor{red}{+0.06})}}$ \\
  & PSNR   & 33.86 & 33.74 & \cellcolor[HTML]{E7ECE4} \underline{34.20}$_{(\textrm{\textcolor{red}{+0.12})}}$ & \cellcolor[HTML]{E7ECE4} 34.11$_{(\textrm{\textcolor{red}{+0.38})}}$ & \cellcolor[HTML]{E7ECE4}\textbf{34.35}$_{(\textrm{\textcolor{red}{+0.32})}}$ \\
  & Params & 31.4 & 102.0 & \cellcolor[HTML]{E7ECE4}\textbf{3.4}$_{(\textrm{\textcolor{red}{-69.91\%})}}$ & \cellcolor[HTML]{E7ECE4}4.5$_{(\textrm{\textcolor{red}{-89.08\%})}}$ & \cellcolor[HTML]{E7ECE4}\underline{3.7}$_{(\textrm{\textcolor{red}{-90.16\%})}}$ \\
  & FLOPs  & 157.0 & 978.0 & \cellcolor[HTML]{E7ECE4}76.6$_{(\textrm{\textcolor{blue}{+2.68\%})}}$ & \cellcolor[HTML]{E7ECE4}\textbf{63.2}$_{(\textrm{\textcolor{red}{-67.92\%})}}$ & \cellcolor[HTML]{E7ECE4}\underline{66.2}$_{(\textrm{\textcolor{red}{-63.63\%})}}$ \\
  \midrule
  \multirow{21}{*}{\rotatebox{90}{WeatherBench}} 
  & & ConvLSTM & PredRNN & PredRNN++ & MIM & E3DLSTM \\
  & MSE & 1.521 & 1.331 & 1.634 & 1.784 & 1.592 \\
  & MAE ($\times 10^{-2}$) & 79.49 & 72.46 & 78.83 & 87.16 & 80.59 \\
  & RMSE & 1.233 & 1.154 & 1.278 & 1.336 & 1.262 \\
  & Params & 15.0 & 23.6 & 38.3 & 37.8 & 51.1 \\
  & FLOPs & 136.0 & 278.0 & 413.0 & 109.0 & 169.0 \\
  \cline{2-7}
  \addlinespace[0.1cm]
  & & PredRNNv2 & MLP-Mixer & \cellcolor{gray!10} Uniformer & \cellcolor{gray!10} SimVP & \cellcolor{gray!10} TAU \\
  & MSE & 1.545 & 1.255 & \cellcolor{gray!10} 1.204 & \cellcolor{gray!10} 1.238 & \cellcolor{gray!10} \underline{1.162} \\
  & MAE ($\times 10^{-2}$) & 79.86 & 70.11 & \cellcolor{gray!10} 68.85 & \cellcolor{gray!10} 70.37 & \cellcolor{gray!10} \underline{67.07} \\
  & RMSE & 1.243 & 1.119 & \cellcolor{gray!10} 1.097 & \cellcolor{gray!10} 1.113 & \cellcolor{gray!10} \underline{1.078} \\
  & Params & 23.6 & 11.1 & \cellcolor{gray!10} 12.0 & \cellcolor{gray!10} 14.7 & \cellcolor{gray!10} 12.2 \\
  & FLOPs & 279.0 & 5.9 & \cellcolor{gray!10} 7.5 & \cellcolor{gray!10} 8.0 & \cellcolor{gray!10} 6.7 \\
  \cline{2-7}
  \addlinespace[0.1cm]
  &     & ConvNeXt & IAM4VP & \cellcolor[HTML]{E7ECE4}USTEP w/Uniformer & \cellcolor[HTML]{E7ECE4}USTEP w/SimVP & \cellcolor[HTML]{E7ECE4}USTEP w/TAU\\
  & MSE & 1.277 & 1.988 & \cellcolor[HTML]{E7ECE4} 1.190$_{(\textrm{\textcolor{red}{-0.014})}}$ & \cellcolor[HTML]{E7ECE4} 1.215$_{(\textrm{\textcolor{red}{-0.023})}}$ &  \cellcolor[HTML]{E7ECE4}\textbf{1.150}$_{(\textrm{\textcolor{red}{-0.012})}}$ \\
  & MAE ($\times 10^{-2}$) & 72.20 & 96.43 & \cellcolor[HTML]{E7ECE4} 67.32$_{(\textrm{\textcolor{red}{-1.53})}}$ & \cellcolor[HTML]{E7ECE4} 69.11$_{(\textrm{\textcolor{red}{-1.26})}}$ & \cellcolor[HTML]{E7ECE4}\textbf{65.83}$_{(\textrm{\textcolor{red}{-1.24})}}$ \\
  & RMSE ($\times 10^{-2}$) & 1.130 & 1.410 & \cellcolor[HTML]{E7ECE4} 1.087$_{(\textrm{\textcolor{red}{-0.010})}}$ & \cellcolor[HTML]{E7ECE4} 1.102$_{(\textrm{\textcolor{red}{-0.011})}}$ & \cellcolor[HTML]{E7ECE4}\textbf{1.072}$_{(\textrm{\textcolor{red}{-0.006})}}$ \\
  & Params   & 10.1 & 12.9 & \cellcolor[HTML]{E7ECE4}\textbf{3.3}$_{(\textrm{\textcolor{red}{-72.50\%})}}$ & \cellcolor[HTML]{E7ECE4}4.4$_{(\textrm{\textcolor{red}{-70.07\%})}}$ & \cellcolor[HTML]{E7ECE4}\underline{3.6}$_{(\textrm{\textcolor{red}{-70.49\%})}}$ \\
  & FLOPs & \underline{5.7} & \textbf{2.4} & \cellcolor[HTML]{E7ECE4}7.7$_{(\textrm{\textcolor{blue}{+2.67\%})}}$ & \cellcolor[HTML]{E7ECE4}9.9$_{(\textrm{\textcolor{blue}{+23.75\%})}}$ & \cellcolor[HTML]{E7ECE4}8.2$_{(\textrm{\textcolor{blue}{+22.39\%})}}$ \\
  \bottomrule
  \end{tabular}
  }
  \label{tab:equal}
\end{table*}

\begin{figure*}[ht]
  \centering
  \includegraphics[width=1.0\textwidth]{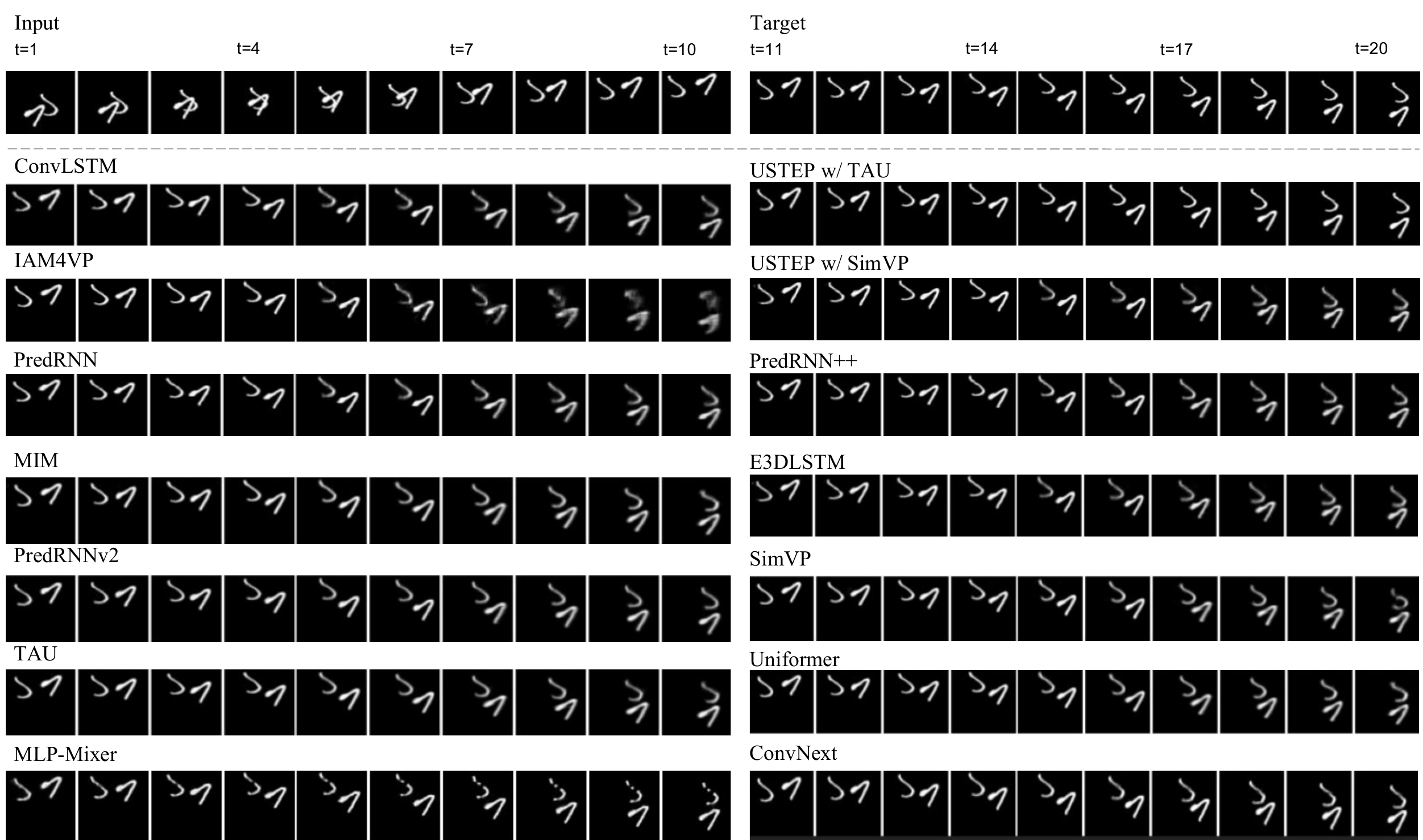}
  \caption{The qualitative visualization on Moving MNIST. }
  \label{fig:moving_mnist_main}
  \vspace{-4mm}
\end{figure*}

\subsection{Equal Frame Task}

Under the experimental setup of the Equal frame task paradigm, we meticulously conducted a comprehensive evaluation of the model's performance on three distinct datasets. These datasets, namely Moving MNIST~\cite{srivastava2015unsupervised}, Human3.6M~\cite{ionescu2013human3}, and WeatherBench~\cite{rasp2020weatherbench}, were judiciously chosen for their inherent property of symmetrical frame sequences, where the number of output frames precisely mirrors that of the input frames. This characteristic facilitates a rigorous assessment of the model's capacity to generate temporally coherent and contextually accurate predictions over varying domains and complexity levels.

The Moving MNIST dataset requires the model to forecast 10 subsequent frames based on an equivalent number of preceding frames. This task challenges the model to understand and predict the dynamics of two moving digits within a frame, encapsulating the complexity of motion patterns and interactions. Similarly, the Human3.6M dataset, derived from a rich repository of human activities, mandates the prediction of 4 future frames from 4 given frames. This dataset serves as a crucible for evaluating the model's proficiency in capturing and forecasting human motion dynamics. The WeatherBench dataset introduces a unique challenge: predicting 12 future temperature patterns from 12 preceding snapshots. This scenario tests the model's ability to grasp and anticipate complex meteorological evolutions over time. The detailed results are summarized in Table~\ref{tab:equal}. 

\begin{figure*}[ht]
\centering
\vspace{2mm}
\includegraphics[width=1.0\textwidth]{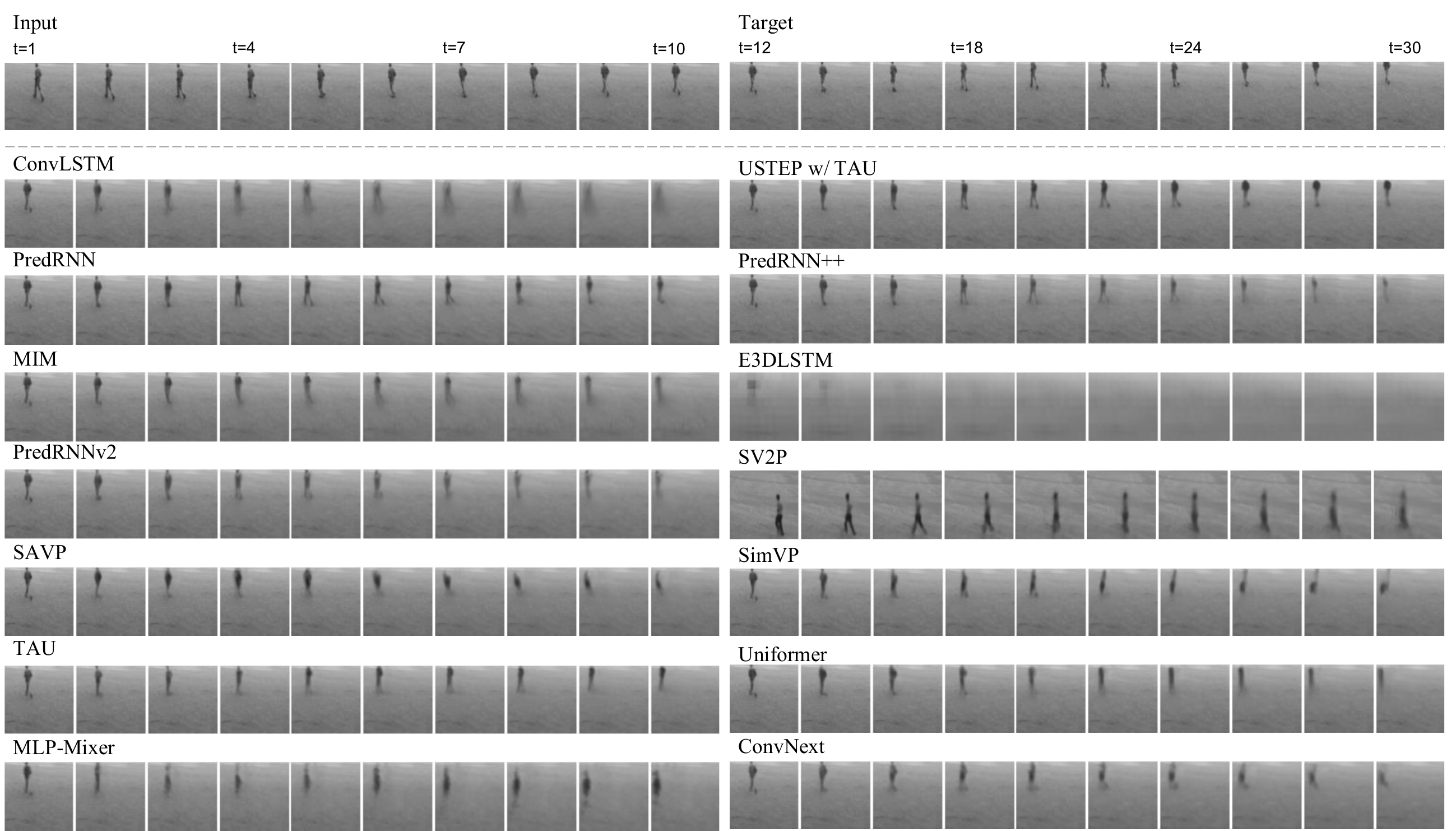}
\caption{The qualitative visualization on KTH. }
\vspace{-4mm}
\label{fig:kth_visual_main}
\end{figure*}

In the Moving MNIST dataset, USTEP w/TAU outperforms the previously top-ranked recurrent-based model, PredRNN++, not only in predictive accuracy but also in computational efficiency. Remarkably, USTEP achieves these superior results while requiring only half the number of parameters and a mere tenth of the FLOPs compared to PredRNN++. In a direct comparison with the leading recurrent-free model, TAU, USTEP not only operates within a comparable range of FLOPs but does so with approximately 42\% of TAU's parameters, thereby demonstrating significant improvements across various performance metrics. This leap in efficiency and effectiveness is illustrated through a compelling visualization example presented in Fig.~\ref{fig:moving_mnist_main}, highlighting USTEP's prowess in generating highly accurate predictions in the Moving MNIST challenge.

On the Human3.6M, a complex dataset capturing a wide array of human activities, USTEP w/TAU continues to demonstrate superior performance. Notably, it employs one-third of the parameters required by its closest competitor, Uniformer, and operates with approximately 12\% fewer FLOPs. Moreover, USTEP w/Uniformer stands out as the leading model in all evaluated metrics with about 30\% fewer parameters. This significant reduction in computational resources, without a compromise in performance, underscores USTEP's capability to predict human motion with remarkable efficiency.

The assessment on WeatherBench, a challenging dataset for meteorological prediction, further solidifies USTEP's position as a versatile and potent model in the arena of spatiotemporal forecasting. USTEP w/TAU obtains impressive advancements over existing models across a spectrum of metrics. It maintains FLOPs close to the premier recurrent-free model, TAU, while its parameter count is merely 29\% of TAU's. Furthermore, when juxtaposed with the leading recurrent-based model, PredRNN, USTEP's parameter efficiency becomes even more pronounced, with its parameters constituting a mere 15\% of those required by PredRNN. Among these datasets, USTEP demonstrates significant improvements over TAU, SimVP, and Uniformer in performance metrics. Notably, it achieves these enhancements with a lower parameter count and similar FLOPs, showcasing its robustness and versatility across diverse domains. Through these rigorous evaluations, USTEP showcases its prowess in precise and efficient prediction across various datasets under the equal frame task setting.

\subsection{Extended Frame Task}

\begin{table*}[h]
  \centering
  \caption{The quantitative results of different methods on the KTH dataset ($10 \rightarrow 20$ frames). The results of the USTEP series that underperform the default are in \textcolor{blue}{blue} while outperforming those are in \textcolor{red}{red}.}
  \small
  \renewcommand\arraystretch{1.25}
  \setlength{\tabcolsep}{1.4mm}{
    \begin{tabular}{cccccccccc}
      \toprule
      \multicolumn{1}{c}{Method} & Param & FLOPs & MSE $\downarrow$ & MAE $\downarrow$ & SSIM $\uparrow$ & PSNR $\uparrow$ & LPIPS $\downarrow$ & FVD $\downarrow$ \\
      \midrule
      ConvLSTM   & 14.9   & 1368.0 & 47.65  & 445.5 & 0.8977 & 26.99 & 26.69 & 640.78 \\
      PredRNN    & 23.6   & 2800.0 & 41.07  & 380.6 & 0.9097 & 27.95 & 21.89 & 436.26 \\
      PredRNN++  & 38.3   & 4162.0 & 39.84  & 370.4 & 0.9124 & 28.13 & \underline{19.87} & \underline{394.63} \\
      MIM        & 39.8   & 1099.0 & 40.73  & 380.8 & 0.9025 & 27.78 & \textbf{18.81} & 1105.97 \\
      E3DLSTM    & 53.5   & 217.0  & 136.40 & 892.7 & 0.8153 & 21.78 & 48.36 & 1998.82 \\
      PredRNNv2  & 23.6  & 2815.0 & 39.57  & \underline{368.8} & 0.9099 & 28.01 & 21.48 & 500.30 \\
      SV2P      & 8.3  & 468.0 & 64.88 & 687.2 & 0.8524 & 27.77 & 27.60 & 433.19 \\
      SAVP      & 17.6 & 641.0 & 124.53 & 867.8 & 0.8192 & 26.00 & 23.41 & 685.70 \\
      \hline
      MLP-Mixer  & 20.3   & 66.6   & 57.74  & 517.4 & 0.8886 & 25.72 & 28.80 & 761.58 \\
      ConvNeXt   & 12.5   & \underline{63.9}   & 45.48  & 428.3 & 0.9037 & 26.96 & 26.25 & 569.33 \\
      \rowcolor{gray!10}  Uniformer  & \textbf{11.8}   & 78.3   & 44.71  & 404.6 & 0.9058 & 27.16 & 24.17 & 543.40 \\
      \rowcolor{gray!10} SimVP      & 12.2 & \textbf{62.8} & 41.11 & 397.1 & 0.9065 & 27.46 & 26.50 & 567.72 \\
      \rowcolor{gray!10} TAU        & 15.0   & 73.8   & 45.32  & 421.7 & 0.9086 & 27.10 & 22.86 & 530.62 \\
      \hline
      \rowcolor[HTML]{E7ECE4}USTEP w/Uniformer & \underline{12.0}$_{(\textrm{\textcolor{blue}{+1.69\%})}}$ & 101.0$_{(\textrm{\textcolor{blue}{+28.99\%})}}$ & 40.23$_{(\textrm{\textcolor{red}{-4.48})}}$ & 377.5$_{(\textrm{\textcolor{red}{-27.1})}}$ & \underline{0.9140}$_{(\textrm{\textcolor{red}{+0.0082})}}$ & \underline{28.55}$_{(\textrm{\textcolor{red}{+1.39})}}$ & 20.15$_{(\textrm{\textcolor{red}{-4.02})}}$ & 420.30$_{(\textrm{\textcolor{red}{-123.10})}}$ \\
      \rowcolor[HTML]{E7ECE4}USTEP w/SimVP & 16.4$_{(\textrm{\textcolor{blue}{+34.43\%})}}$ & 137.0$_{(\textrm{\textcolor{blue}{+118.15\%})}}$ & \underline{39.75}$_{(\textrm{\textcolor{red}{-1.36})}}$ & 372.4$_{(\textrm{\textcolor{red}{-24.7})}}$ & 0.9135$_{(\textrm{\textcolor{red}{+0.0070})}}$ & 28.47$_{(\textrm{\textcolor{red}{+1.01})}}$ & 20.35$_{(\textrm{\textcolor{red}{-6.15})}}$ & 410.80$_{(\textrm{\textcolor{red}{-156.92})}}$ \\
      \rowcolor[HTML]{E7ECE4}USTEP w/TAU   & 12.8$_{(\textrm{\textcolor{red}{-14.67\%})}}$   & 107.0$_{(\textrm{\textcolor{blue}{+44.99\%})}}$  & \textbf{39.55}$_{(\textrm{\textcolor{red}{-5.77})}}$ & \textbf{364.9}$_{(\textrm{\textcolor{red}{-56.8})}}$ & \textbf{0.9165}$_{(\textrm{\textcolor{red}{+0.0079})}}$ & \textbf{28.98}$_{(\textrm{\textcolor{red}{+1.88})}}$ & 19.96$_{(\textrm{\textcolor{red}{-2.90})}}$ & \textbf{393.13}$_{(\textrm{\textcolor{red}{-137.49})}}$ \\
      \bottomrule
    \end{tabular}%
  }
  \label{tab:kth}
\end{table*}

\begin{table*}[h!]
  \fontsize{8.5pt}{11pt}\selectfont
  \centering
  \caption{Quantitative results on the KTH dataset ($10 \rightarrow 40$ frames). The results of the USTEP series that underperform the default are in \textcolor{blue}{blue} while outperforming those are in \textcolor{red}{red}.}
  \renewcommand\arraystretch{1.25}
  \setlength{\tabcolsep}{1.7mm}{
    \begin{tabular}{ccccccccc}
      \toprule
      \multicolumn{1}{c}{Method} & Params & FLOPs & MSE $\downarrow$ & MAE $\downarrow$ & SSIM $\uparrow$ & PSNR $\uparrow$ & LPIPS $\downarrow$ & FVD $\downarrow$ \\
      \midrule
      ConvLSTM   & 14.9   & 2312.0  & 63.60  & 512.2  &  0.8677  &  25.42  & 36.98 & 1893.45 \\
      PredRNN    & 23.6   & 4730.0  & 61.42  & 499.0  & 0.8753  & 25.49  & 34.26 & 1711.06 \\
      PredRNN++  & 38.3   & 7032.0 & 74.21   & 554.1  & 0.8615 & 24.53  &  38.44 & 1482.15 \\
      MIM        & 39.8   & 1864.0  & 60.92  & 520.3  & 0.8645  & 25.42  & 35.14 & 1578.64 \\
      E3DLSTM    & 53.5  & 498.0 & 192.83 & 1124.5 & 0.7843 & 19.87 & 68.43 & 3995.82 \\
      PredRNNv2  & 23.6   & 4757.0  & 70.27  & 516.7 & 0.8554  & 25.12 & 37.43 & 1500.34 \\
      SV2P      & \textbf{8.3}  & 638.0 & 85.14 & 907.1 & 0.8011 & 24.08 & 38.24 & 1681.09 \\
      SAVP      & 17.6 & 1084.0 & 156.43 & 1198.6 & 0.8042 & 24.32 & 39.90 & 1691.11 \\
      \hline
      MLP-Mixer  & 20.3   & \underline{133.2}  &  81.44  & 681.0  & 0.8397  & 23.36 & 42.55 & 2148.75 \\
      ConvNeXt   & 12.5   & 127.8  & 58.57  & 502.1  & 0.8760  & 25.59  & 36.77 & 1648.97 \\
      \rowcolor{gray!10} Uniformer  & \underline{11.8} & 156.6  & 63.04  & 561.0  & 0.8601  & 24.98 &  39.32 & 1543.23 \\
      \rowcolor{gray!10} SimVP      & 12.2 & \textbf{125.6}  & 68.63  & 603.5  & 0.8555  & 25.10  &  39.23 & 1501.04 \\
      \rowcolor{gray!10} TAU        & 15.0   & 147.6 & 62.22  &  511.1  & 0.8689  &  25.47 & 35.77 & 1500.63 \\
      \hline
      \rowcolor[HTML]{E7ECE4}USTEP w/Uniformer & 12.0$_{(\textrm{\textcolor{blue}{+1.69\%})}}$ & 202.0$_{(\textrm{\textcolor{blue}{+28.99\%})}}$ & \underline{54.84}$_{(\textrm{\textcolor{red}{-8.20})}}$ & \underline{440.5}$_{(\textrm{\textcolor{red}{-120.5})}}$& \underline{0.8798}$_{(\textrm{\textcolor{red}{+0.0197})}}$ & \underline{26.10}$_{(\textrm{\textcolor{red}{+1.12})}}$ & \underline{32.45}$_{(\textrm{\textcolor{red}{-6.87})}}$ & \underline{945.24}$_{(\textrm{\textcolor{red}{-597.99})}}$   \\
      \rowcolor[HTML]{E7ECE4}USTEP w/SimVP & 16.4$_{(\textrm{\textcolor{blue}{+34.43\%})}}$ & 274.0$_{(\textrm{\textcolor{blue}{+118.15\%})}}$ & 55.23$_{(\textrm{\textcolor{red}{-13.40})}}$ & 445.6$_{(\textrm{\textcolor{red}{-157.9})}}$ & 0.8783$_{(\textrm{\textcolor{red}{+0.0228})}}$ & 26.05$_{(\textrm{\textcolor{red}{+0.95})}}$ & 32.81$_{(\textrm{\textcolor{red}{-6.42})}}$  & 1095.72$_{(\textrm{\textcolor{red}{-405.32})}}$  \\
      \rowcolor[HTML]{E7ECE4}USTEP w/TAU   & 12.8$_{(\textrm{\textcolor{red}{-14.67\%})}}$  & 214.0$_{(\textrm{\textcolor{blue}{+44.99\%})}}$  & \textbf{54.68}$_{(\textrm{\textcolor{red}{-7.54})}}$ & \textbf{433.2}$_{(\textrm{\textcolor{red}{-77.9})}}$ & \textbf{0.8832}$_{(\textrm{\textcolor{red}{+0.0143})}}$  & \textbf{26.22}$_{(\textrm{\textcolor{red}{+0.75})}}$ &  \textbf{31.28}$_{(\textrm{\textcolor{red}{-4.49})}}$  & \textbf{849.16}$_{(\textrm{\textcolor{red}{-651.47})}}$  \\
      \bottomrule
    \end{tabular}%
  }
  \label{tab:kth40}
\end{table*}

In the exploration of spatiotemporal predictive models' capabilities, particularly in the context of forecasting a greater number of frames than those observed, the KTH dataset emerges as a pivotal benchmark. The intricacies of this task lie in its demand for models to not only understand and internalize short-term sequences but also to extrapolate these observations into longer, future sequences, thereby amplifying the challenge of accurate prediction.

Upon a detailed examination, as presented in Table~\ref{tab:kth}, a distinct pattern emerges. Recurrent-free models, though commendable for their efficiency in terms of both parameters (Params) and computational complexity (FLOPs), tend to fall short in achieving the high-performance benchmarks set by their recurrent-based counterparts. However, within this competitive landscape, USTEP distinctly outperforms the previously established frontrunner, PredRNN++, both in terms of predictive accuracy and computational efficiency. USTEP not only aligns itself with the efficiency metrics typically seen in recurrent-free models but also surpasses them in performance. This juxtaposition is illustrated in Fig.~\ref{fig:kth_visual_main}, where USTEP w/TAU predicts notably more realistic and sharply defined compared to those of other methods. While TAU, a recurrent-free model, exhibits commendable performance, it struggles to replicate the nuanced details that USTEP w/TAU manages to capture. By integrating Uniformer, SimVP, and TAU as the basic modules, USTEP consistently improves their baseline models across a range of evaluation metrics, showcasing its robustness and versatility in this setting. For example, USTEP w/TAU achieves an SSIM of 0.9165 and an FVD of 393.13, outperforming both its recurrent-free and recurrent-based counterparts.

To push the boundaries of predictive modeling further, we extended the challenge on the KTH dataset to encompass a longer prediction horizon, specifically from 10 observed frames to forecasting 40 future frames. The results of this ambitious undertaking are meticulously compiled in Table~\ref{tab:kth40}. USTEP w/TAU not only maintains but also amplifies its lead over competing models, delivering outstanding performance across a suite of evaluation metrics. This impressive feat underscores USTEP's exceptional capability to generate long sequences of future frames with high fidelity, thereby affirming its robustness and versatility for a broad spectrum of real-world scenarios. In particular, USTEP w/TAU achieves an MSE of 54.68, which is significantly lower than the baseline models, and an SSIM of 0.8832, indicating superior structural similarity in the predicted frames. Furthermore, USTEP w/TAU also demonstrates a notable reduction in FVD scores, highlighting its enhanced ability to produce visually coherent and temporally consistent predictions. These improvements are not limited to one configuration; USTEP variants with Uniformer and SimVP also show considerable improvements, showcasing the framework's flexibility and effectiveness across different underlying architectures.

Such comprehensive evaluations serve not only to benchmark the current state of spatiotemporal predictive modeling but also to illuminate the path forward. While traditional recurrent-based methods like PredRNN struggle to maintain performance over longer prediction horizons, and recurrent-free methods like SimVP tend to overlook short-term dependencies, USTEP manages to harness the strengths of both, setting a new standard in scenarios demanding the prediction of extended future sequences.

\subsection{Reduced Frame Task}

\begin{table*}[ht]
  \fontsize{8.5pt}{11pt}\selectfont
  \centering
  \caption{Quantitative results on the Caltech Pedestrian dataset ($10 \rightarrow 1$ frame). The results of the USTEP series that underperform the default are in \textcolor{blue}{blue} while outperforming those are in \textcolor{red}{red}.}
  \vspace{-2mm}
  \renewcommand\arraystretch{1.25}
  \setlength{\tabcolsep}{3.3mm}{
  \begin{tabular}{cccccccc}
  \toprule
  Method    & Params & FLOPs & MSE $\downarrow$ & MAE $\downarrow$ & SSIM $\uparrow$ & PSNR $\uparrow$ & LPIPS $\downarrow$\\
  \midrule
  ConvLSTM       & 15.0  & 595.0  & 139.6  & 1583.3 & 0.9345 & 27.46 & 8.58 \\
  PredRNN        & 23.7  & 1216.0 & 130.4 & 1525.5 & 0.9374 & 27.81 & 7.40 \\
  PredRNN++      & 38.5 & 1803.0 & 125.5  & 1453.2 & 0.9433 & 28.02 & 13.21 \\
  MIM            & 49.2  & 1858.0 & 125.1 & 1464.0  & 0.9409 & 28.10 & 6.35 \\
  E3DLSTM        & 54.9  & 1004.0 & 200.6  & 1946.2  & 0.9047 & 25.45 & 12.60 \\
  PredRNNv2      & 23.8  & 1223.0 & 147.8  & 1610.5 & 0.9330 & 27.12 & 8.92 \\
  \hline
  MLP-Mixer      & 22.2  & 83.5  & 207.9  & 1835.9  & 0.9133 & 26.29 & 7.75 \\
  ConvNeXt       & 12.5  & 80.2  & 146.8  & 1630.0 & 0.9336 & 27.19 & 6.99 \\
  \rowcolor{gray!10} Uniformer      & 11.8  & 104.0  & 135.9  & 1534.2 & 0.9393 & 27.66 & 6.87 \\
  \rowcolor{gray!10} SimVP          & 8.6 & \textbf{60.6} & 160.2 & 1690.8 & 0.9338 & 26.81 & 6.76 \\
  \rowcolor{gray!10} TAU            & 15.0 & 92.5 & 131.1 & 1507.8 & 0.9456 & 27.83 & \underline{5.49} \\
  \hline
  \rowcolor[HTML]{E7ECE4}USTEP w/Uniformer & \textbf{4.7}$_{(\textrm{\textcolor{red}{-60.17\%})}}$ & \underline{79.6}$_{(\textrm{\textcolor{red}{-23.46\%})}}$ & \underline{123.8}$_{(\textrm{\textcolor{red}{-12.1})}}$ & \underline{1421.2}$_{(\textrm{\textcolor{red}{-113.0})}}$  & \underline{0.9471}$_{(\textrm{\textcolor{red}{+0.0078})}}$  & \underline{28.32}$_{(\textrm{\textcolor{red}{+0.66})}}$  & 5.50$_{(\textrm{\textcolor{red}{-1.37})}}$ \\
  \rowcolor[HTML]{E7ECE4}USTEP w/SimVP & 6.3$_{(\textrm{\textcolor{red}{-26.74\%})}}$ & 105.0$_{(\textrm{\textcolor{blue}{+73.27\%})}}$ & 144.6$_{(\textrm{\textcolor{red}{-15.6})}}$ & 1532.0$_{(\textrm{\textcolor{red}{-158.8})}}$  & 0.9465$_{(\textrm{\textcolor{red}{+0.0127})}}$  & 28.09$_{(\textrm{\textcolor{red}{+1.28})}}$ & 5.65$_{(\textrm{\textcolor{red}{-1.11})}}$ \\
  \rowcolor[HTML]{E7ECE4}USTEP w/TAU & \underline{5.1}$_{(\textrm{\textcolor{red}{-66.00\%})}}$  & 85.3$_{(\textrm{\textcolor{red}{-7.78\%})}}$ & \textbf{123.6}$_{(\textrm{\textcolor{red}{-7.5})}}$  & \textbf{1407.9}$_{(\textrm{\textcolor{red}{-99.9})}}$  &\textbf{0.9477}$_{(\textrm{\textcolor{red}{+0.0021})}}$  & \textbf{28.37}$_{(\textrm{\textcolor{red}{+0.54})}}$ & \textbf{4.94}$_{(\textrm{\textcolor{red}{-0.55})}}$ \\
  \bottomrule
  \end{tabular}
  }
  \label{tab:caltech}
\vspace{-3mm}
\end{table*}

\begin{table*}[ht]
  \fontsize{8.5pt}{11pt}\selectfont
  \centering
  \caption{Quantitative results on the SEVIR dataset ($13 \rightarrow 12$ frames) and UCF Sports ($4 \rightarrow 1$ frame). The results of the USTEP series that underperform the default are in \textcolor{blue}{blue} while outperforming those are in \textcolor{red}{red}.}
  \vspace{-2mm}
  \renewcommand\arraystretch{1.25}
  \setlength{\tabcolsep}{2.0mm}{
  \begin{tabular}{ccccccccc}
  \toprule
  & \multicolumn{4}{c}{SEVIR} & \multicolumn{4}{c}{UCF Sports} \\ \cline{2-5} \cline{6-9} 
  Method  & Params & FLOPs & CSI$\uparrow$ & RMSE$\downarrow$ & Params & FLOPs & PSNR$\uparrow$ & LPIPS$\downarrow$ \\
  \midrule
  ConvLSTM       & 17.4 & 3097.0 & 0.4082 & 13.01 & 15.0 & 4017.0  & 25.90 & 29.33 \\
  PredRNN        & 27.9 & 6170.0 & 0.4288 & 12.82 & 23.8 & 8208.0  & 27.17 & 28.15 \\
  PredRNN++      & 42.6 & 8706.0 & \underline{0.4312} & 12.64 & 38.5 & 12171.0 & 27.26 & 26.80 \\
  \hline
  \rowcolor{gray!10} Uniformer & 19.6 & 408.0 & 0.3797 & 13.87 & 38.7 & \textbf{223.0} & 26.63 & 28.78 \\
  \rowcolor{gray!10} SimVP     & \underline{12.8} & \textbf{171.0} & 0.3959 & 12.66 & 37.9 & \underline{226.0} & 27.96 & 24.99 \\
  \rowcolor{gray!10} TAU       & 20.5 & \underline{241.0} & 0.3941 & 12.73 & 38.2  & 236.0 & \underline{28.10} & \underline{24.55} \\
  \hline
  \rowcolor[HTML]{E7ECE4}USTEP w/Uniformer & \textbf{12.2}$_{(\textrm{\textcolor{red}{-37.75\%})}}$ & 253.0$_{(\textrm{\textcolor{red}{-37.99\%})}}$ & 0.4091$_{(\textrm{\textcolor{red}{+0.0294})}}$ & 13.06$_{(\textrm{\textcolor{red}{-0.81})}}$  & \textbf{5.6}$_{(\textrm{\textcolor{red}{-85.53\%})}}$  & 235.0$_{(\textrm{\textcolor{blue}{+5.38\%})}}$ & 27.32$_{(\textrm{\textcolor{red}{+0.69})}}$ & 26.91$_{(\textrm{\textcolor{red}{-1.87})}}$ \\
  \rowcolor[HTML]{E7ECE4}USTEP w/SimVP & 13.1$_{(\textrm{\textcolor{blue}{+2.34\%})}}$ & 267.0$_{(\textrm{\textcolor{blue}{+56.14\%})}}$ & 0.4289$_{(\textrm{\textcolor{red}{+0.0330})}}$ & \underline{12.62}$_{(\textrm{\textcolor{red}{-0.04})}}$ & \underline{6.1}$_{(\textrm{\textcolor{red}{-83.91\%})}}$ & 251.0$_{(\textrm{\textcolor{blue}{+11.06\%})}}$ & 28.01$_{(\textrm{\textcolor{red}{+0.05})}}$ & 24.77$_{(\textrm{\textcolor{red}{-0.22})}}$ \\
  \rowcolor[HTML]{E7ECE4}USTEP w/TAU & 13.5$_{(\textrm{\textcolor{red}{-34.15\%})}}$ & 277.0$_{(\textrm{\textcolor{blue}{+14.93\%})}}$ & \textbf{0.4366}$_{(\textrm{\textcolor{red}{+0.0425})}}$  & \textbf{12.10}$_{(\textrm{\textcolor{red}{-0.63})}}$  & 6.3$_{(\textrm{\textcolor{red}{-83.51\%})}}$ & 262.0$_{(\textrm{\textcolor{blue}{+11.12\%})}}$ & \textbf{28.48}$_{(\textrm{\textcolor{red}{+0.38})}}$ & \textbf{24.23}$_{(\textrm{\textcolor{red}{-0.32})}}$ \\
  \bottomrule
  \end{tabular}
  }
  \label{tab:sevir}
\vspace{-3mm}
\end{table*}

Within the scope of the Reduced Frame Task, our objective was to rigorously evaluate the model's adeptness at learning from a limited set of observed frames and its effectiveness in minimizing accumulated prediction errors. This evaluation is critical, as it tests the model's capability to infer future states from sparse temporal data, a common scenario in real-world applications. The Caltech Pedestrian dataset, known for its complexity and real-life relevance, served as the foundation for this assessment. The dataset's diverse and dynamic pedestrian scenes provide a robust testbed for spatio-temporal prediction models. The results from this evaluation are systematically presented in Table~\ref{tab:caltech}. 

Recurrent-based methods such as PredRNN and PredRNN++ exhibit strong overall performance with balanced metrics, indicating their proficiency in handling temporal dependencies. However, these methods tend to have higher computational costs in terms of parameters and FLOPs. Recurrent-free methods like SimVP and TAU demonstrate low computational costs but at the expense of higher prediction errors, suggesting limitations in capturing detailed temporal dependencies. In contrast, USTEP w/TAU delivers the best overall performance, achieving the lowest MSE and MAE, highest SSIM and PSNR, and lowest LPIPS, all while significantly reducing computational requirements. This highlights USTEP's ability to balance accuracy and efficiency effectively. Additionally, USTEP variants with Uniformer and SimVP also exhibit notable improvements over their baseline models, demonstrating significant reductions in parameters and FLOPs while maintaining or enhancing performance metrics. These results underscore the efficacy of the USTEP framework in integrating the strengths of both recurrent-based and recurrent-free methods.

The SEVIR dataset is particularly challenging due to the highly dynamic and irregular nature of meteorological events. As shown in Table~\ref{tab:sevir}, recurrent-based methods like PredRNN++ achieved strong results, with the highest Critical Success Index (CSI) among non-USTEP models, but at high computational overhead cost. Recurrent-free methods such as SimVP obtained exceptional computational efficiency, with the lowest parameter and FLOP counts, but lagged behind in CSI. The TAU model provided a balance between accuracy and efficiency, achieving competitive results. The USTEP framework outperformed baselines on the SEVIR dataset. USTEP w/TAU delivered the best performance, achieving the highest CSI and the lowest RMSE, while reducing parameters by 34.15\%. The UCF Sports dataset was used to evaluate the model's ability to predict human actions in dynamic scenes. USTEP demonstrated its adaptability and superior performance on this dataset. USTEP w/TAU achieved the best overall results, with the highest PSNR and the lowest LPIPS, while reducing parameters by 83.51\% compared to recurrent-based methods.

\begin{table*}[h!]
  \centering
  \caption{Ablation study on the influence of different design choices on the Moving MNIST dataset. ($\Delta T$ is default to 10). Training time is measured in seconds per epoch, while inference time is measured in seconds of the test set.}
  \vspace{-2mm}
  \renewcommand\arraystretch{1.25}
  \setlength{\tabcolsep}{3.6mm}{
  \begin{tabular}{l ccccccccc}
  \toprule
  & \multicolumn{6}{c}{Moving MNIST} \\
  \multirow{1}{*}{Method} & Params & FLOPs & Training time & Inference time  & MSE$\downarrow$ & MAE$\downarrow$ & SSIM$\uparrow$ & PSNR$\uparrow$ \\
  \midrule
  USTEP ($\Delta t = 5$) & 18.9 & 17.7 & 93 & 58 & \textbf{21.84} & 63.21& \textbf{0.9538}& 24.06\\
  \hline
  USTEP ($\Delta t = 1$)  & 18.8 & 52.5 & 600 & 1052 & 31.94 & 71.72 &0.9416 &23.36 \\
  USTEP ($\Delta t = 2$)  & 18.3 & 30.8 & 282 & 293 & 24.62 & \textbf{62.30} & 0.9525 & \textbf{24.46} \\
  USTEP ($\Delta t = 10$) & 19.1 & 13.4 & 64 & 35 & 25.13 & 74.31 & 0.9440 & 23.02 \\
  USTEP ($\Delta t = 1, \Delta T = 1$) & 18.5 & 89.8 & 863 & 1449 & 37.21 & 86.51 & 0.9240 & 22.21 \\
  w/o cross segment       & 17.4 & 13.1 & 82 & 52 & 24.01 & 67.65 & 0.9489 & 23.57 \\
  \hline
  TAU   & 44.7 & 16.0 & 63 & 39 & 24.60 & 71.93 & 0.9454 & 23.19 \\
  PredRNN++ & 38.6 & 171.7 & 511 & 373 & 22.06 & 69.58 & 0.9509 & 23.65 \\
  \bottomrule
  \end{tabular}}
  \label{tab:ablation}
\end{table*}

\subsection{Ablation Study}

To unravel the nuances that underpin the performance of USTEP, an in-depth ablation study was undertaken. This meticulous examination was directed towards understanding the repercussions of various design decisions, with a particular focus on the temporal stride ($\Delta t$) and the integration of a cross-segment mechanism within the model's architecture. For convenience, we refer to USTEP w/TAU as simply USTEP. The findings from this study succinctly encapsulated in Table~\ref{tab:ablation}, shed light on the intricate balance between computational efficiency and predictive accuracy.

The optimal performance is achieved when $\Delta t$ is set to 5, where USTEP records an MSE of 21.84 and an SSIM of 0.9538, marking the zenith of its predictive accuracy. This particular configuration of $\Delta t$ underscores a harmonious balance between capturing essential temporal dynamics and maintaining computational pragmatism. With considerable performance improvements over TAU, USTEP only marginally sacrifices training and inference time while still remaining significantly faster than its recurrent-based counterparts. However, diminishing the stride ($\Delta t$) heightens the model's computational demand, evidenced by an uptick in FLOPs, attributed to the increment in computational occurrences necessitated by shorter temporal intervals. 

The role of the cross-segment mechanism emerges as pivotal within this contextual framework. Its presence is instrumental, acting as a crucial part of USTEP's architecture that significantly enhances spatiotemporal predictive performance. Disabling this cross-segment mechanism results in noticeable performance degradation, emphasizing its criticality in facilitating a nuanced interplay between different temporal segments, thereby enriching the model's predictive depth and accuracy.

\begin{table*}[h!]
  \centering
  \caption{Empirical $\Delta t$ effects. The results that underperform the default are in \textcolor{blue}{blue} while outperforming those are in \textcolor{red}{red}.}
  \renewcommand\arraystretch{1.25}
  \setlength{\tabcolsep}{6.5mm}{
  \begin{tabular}{c lllllll}
  \toprule
  Dataset & $\Delta t$ & Length & MSE$\downarrow$ & MAE$\downarrow$ & SSIM$_{(\times 10^{-2})}$ $\uparrow$ & PSNR$\uparrow$ \\
  \midrule
  MMNIST & 5 & 20 & 21.84$_{(\textrm{\textcolor{black}{+0.00})}}$ & 63.21$_{(\textrm{\textcolor{black}{+0.00})}}$& 95.38$_{(\textrm{\textcolor{black}{+0.00})}}$& 24.06$_{(\textrm{\textcolor{black}{+0.00})}}$ \\
  KTH  & 5 & 30 & 39.91$_{(\textrm{\textcolor{blue}{+0.36})}}$ & 355.3$_{(\textrm{\textcolor{red}{-9.6})}}$ & 91.55$_{(\textrm{\textcolor{blue}{-0.10})}}$ &28.99$_{(\textrm{\textcolor{red}{+0.01})}}$ \\
  Caltech  & 5 & 11  & 123.6$_{(\textrm{\textcolor{black}{+0.00})}}$ & 1407.9$_{(\textrm{\textcolor{black}{+0.00})}}$ & 94.77$_{(\textrm{\textcolor{black}{+0.00})}}$ & 28.37$_{(\textrm{\textcolor{black}{+0.00})}}$ \\
  WeatherBench & 5 & 24 & 1.14$_{(\textrm{\textcolor{red}{-0.01})}}$ & 0.69$_{(\textrm{\textcolor{blue}{+0.04})}}$ & - &  - \\
  Human & 2 & 8 & 109.5$_{(\textrm{\textcolor{black}{+0.00})}}$ & 1380.5$_{(\textrm{\textcolor{black}{+0.00})}}$ & 98.45$_{(\textrm{\textcolor{black}{+0.00})}}$ & 34.35$_{(\textrm{\textcolor{black}{+0.00})}}$ \\
  \bottomrule
  \end{tabular}}
  \label{tab:emprical}
\end{table*}

\begin{figure}[h]
\centering
\vspace{-2mm}
\includegraphics[width=0.98\linewidth]{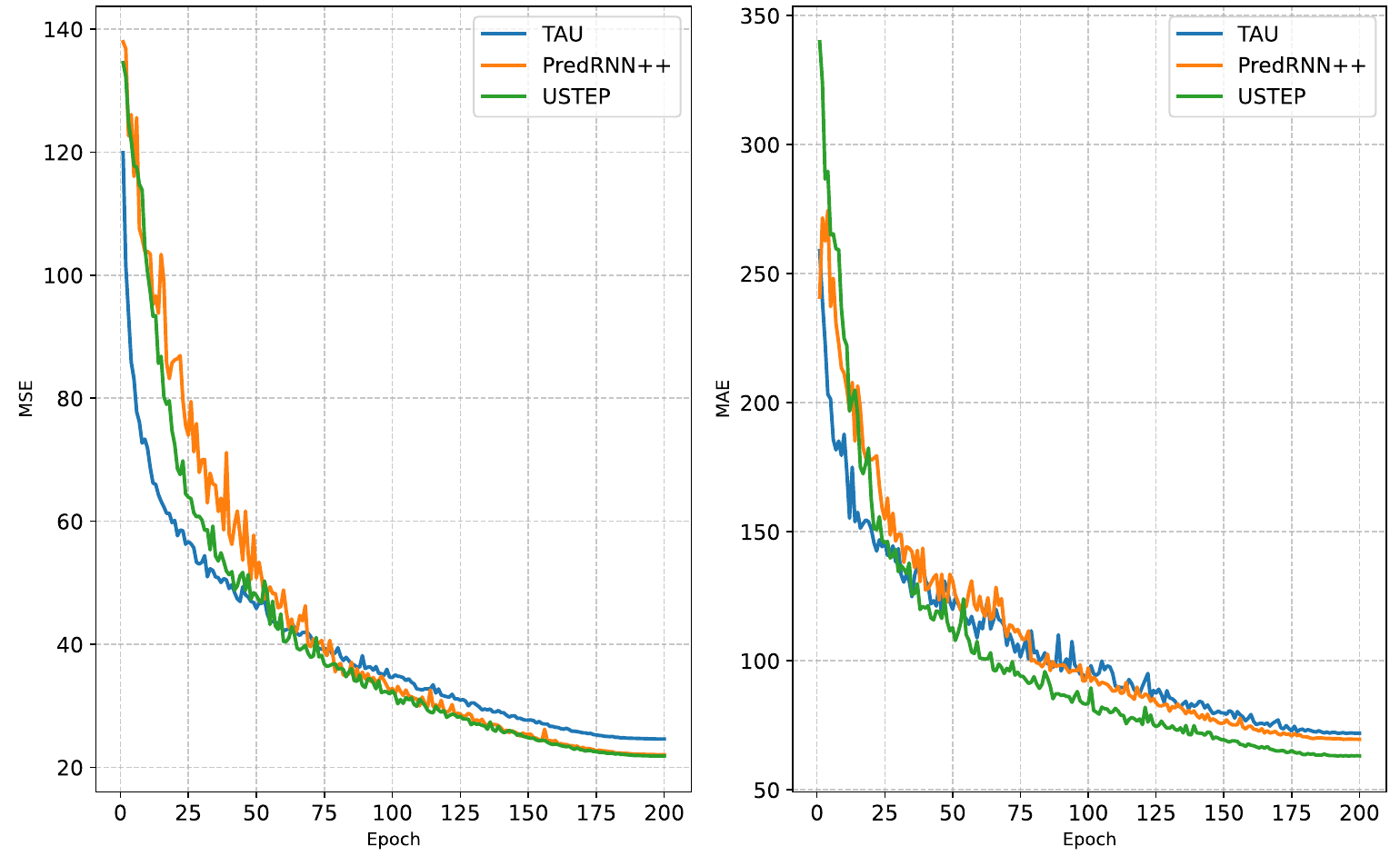}
\caption{Convergence comparison between USTEP and strong baselines, PredRNN++ (recurrent-based) and TAU (recurrent-free).}
\vspace{-4mm}
\label{fig:convergence}
\end{figure}

We compare the convergence curves between USTEP and representative recurrent-based method PredRNN++ and recurrent-free method TAU in Fig.~\ref{fig:convergence}. While TAU benefits from fast convergence in the earlier epochs, USTEP gradually catches up and eventually surpasses TAU. PredRNN++ exhibits the slowest convergence compared to USTEP. Moreover, while USTEP achieves a similar MSE to PredRNN++, it significantly outperforms PredRNN++ in terms of MAE. This suggests that USTEP produces more accurate and visually coherent predicted frames, effectively reducing larger prediction errors. USTEP not only offers a compatible solution that bridges the gap between recurrent-based and recurrent-free models but also provides strong overall performance, making it a robust choice for spatio-temporal predictive learning tasks.
  
\begin{figure}[h]
\centering
\vspace{-2mm}
\includegraphics[width=0.5\textwidth]{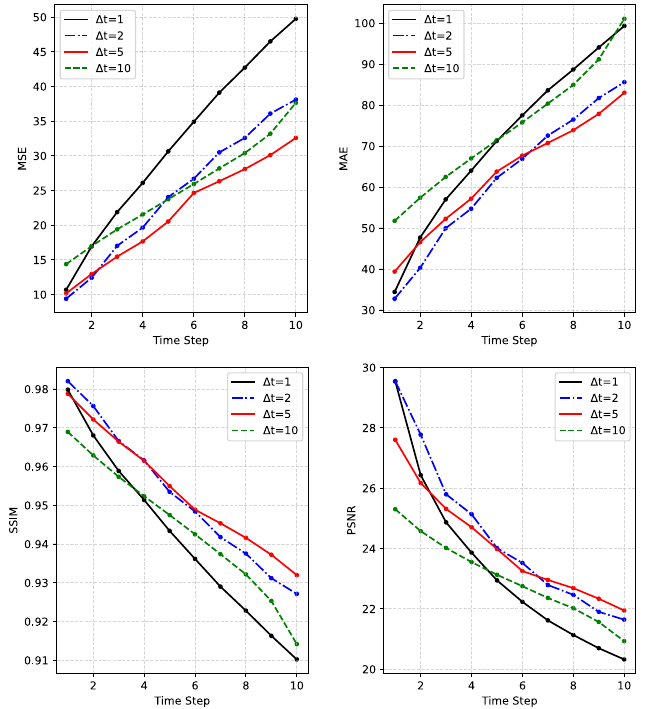}
\caption{Frame-wise comparison in MSE, MAE, SSIM and PSNR metrics. For MSE and MAE metrics, lower values are preferable. For SSIM and PSNR metrics, higher values are more desirable.}
\label{fig:mse}
\vspace{-4mm}
\end{figure}

Fig.~\ref{fig:mse} delineates that $\Delta t = 1$ tends to overemphasize local information, potentially leading to a lack of holistic understanding. In contrast, $\Delta t = 10$ appears to overly prioritize global information, possibly at the expense of missing finer, localized details. These insights underline the importance of choosing an appropriate $\Delta t$ in USTEP to balance local and global temporal considerations, ensuring the holistic integrity of the learned features and predictions.

\subsection{Empirical Analysis and Practical Guideline}

Our objective is to integrate the strengths of both recurrent-based and -free approaches for spatiotemporal predictive learning. Using a fixed time step $\Delta t$ is a simple yet effective way to achieve this. As we have set $\Delta T$ as the length of the input sequence in Definition 3.3, the only hyperparameter left to be tuned is $\Delta t$. Empirically, we recommend using $\Delta t = 5$ for long videos ($T + T' > 10$) and $\Delta t = 2$ for short videos ($T + T' \leq 10$). 
As indicated in Table~\ref{tab:emprical}, minor changes were observed on KTH and WeatherBench. The other datasets remained unchanged in terms of metrics since their optimal $\Delta t$ values were already aligned with our principle. In essence, our empirical principle serves as a foundational guideline for selecting the temporal stride $\Delta t$ when processing video data of varying lengths, guarantees good performance across a wide range of applications.

In practice, $\Delta t$ can also be adjusted based on the frame rate of the video data. For low frame rates (e.g., 1-5 FPS), a smaller $\Delta t$ is preferable to ensure no critical temporal information is skipped. For higher frame rates (e.g., 30 FPS or more), a larger $\Delta t$ can help improve computational efficiency while maintaining performance. Our empirical findings suggest that setting $\Delta t = 5$ for long videos and $\Delta t = 2$ for short videos provides a default configuration.

\subsection{Analysis between USTEP and Representative Baselines}

\noindent\textbf{{Recurrent-based models}} like PredRNN, PredRNN++, and ConvLSTM excel at modeling temporal dependencies by processing frames sequentially, but they suffer from the following limitations:
\begin{itemize}
  \item Redundant Frame-by-Frame Processing: Frame-by-frame processing leads to significant computational overhead, as short-term spatio-temporal redundancies are repeatedly processed. This inefficiency becomes even more pronounced with longer input sequences. For example, in Table~\ref{tab:equal} (Moving MNIST), the FLOPs of ConvLSTM, PredRNN, and PredRNN++ are 56.8, 116.0, and 171.7 GFLOPs, respectively. This shows how the increased complexity of recurrent-based models results in higher computational costs, despite moderate improvements in performance. PredRNN++ achieves better MSE (22.06) compared to PredRNN (23.97) and ConvLSTM (29.80), but the improvement comes at a significant cost in terms of computational efficiency.
  \item Complex Module Design: PredRNN introduces spatio-temporal memory flow, enhancing its ability to capture transitions between frames. PredRNN++ further builds on this by designing more complicated recurrent modules, but the added complexity doesn't always translate to better performance. For instance, in Table~\ref{tab:kth40}, PredRNN++'s FLOPs (7032 GFLOPs) are much higher than ConvLSTM (2312 GFLOPs) and PredRNN (4730 GFLOPs). However, its MSE (74.21) is worse than PredRNN (61.42) and even ConvLSTM in some cases. This suggests that the additional complexity in PredRNN++ does not always yield better results, particularly for long-term predictions.
\end{itemize}

\noindent\textbf{{Recurrent-free models}} such as SimVP and TAU offer an efficient alternative by processing sequences in parallel. However, their limitations include:
\begin{itemize}
  \item Loss of Fine-Grained Temporal Interactions: SimVP processes sequences holistically, focusing on global patterns but neglecting short-term temporal interactions between consecutive frames. This makes it less effective for tasks requiring precise frame-by-frame modeling. For instance, on Moving MNIST (Table~\ref{tab:equal}), SimVP achieves an MSE of 32.15, which is significantly worse than PredRNN (23.97) and PredRNN++ (22.06). Its performance gap demonstrates its inability to capture fine-grained dependencies.
  \item Efficiency vs. Performance Trade-off: TAU simplifies SimVP by replacing the middle U-Net with plain temporal attention units. TAU demonstrates exceptional efficiency compared to recurrent-based models. For example, on Table~\ref{tab:kth40}, TAU achieves a comparable MSE (62.22) with only 147 GFLOPs, whereas PredRNN requires 4730 GFLOPs, and PredRNN++ requires 7032 GFLOPs. However, TAU's focus on global context still limits its ability to model short-term dynamics as effectively as recurrent-based models. While TAU strikes a better balance, it still prioritizes long-term global modeling at the expense of short-term interactions. 
\end{itemize}

\noindent\textbf{{USTEP}} combines the strengths of both recurrent-based and recurrent-free models while addressing their individual limitations. It achieves this through its hierarchical dual-scale framework, which integrates micro-temporal (short-term) and macro-temporal (long-term) modeling:
\begin{itemize}
  \item Avoids Redundant Frame-by-Frame Processing and Captures Fine-Grained Temporal Interactions: Unlike recurrent-based methods, USTEP processes long-term dependencies using macro-temporal segments in a recurrent-free manner, reducing computational overhead while maintaining global context. By incorporating micro-temporal segments, USTEP explicitly models short-term dependencies that recurrent-free models like SimVP and TAU neglect.
  \item Dynamic Integration of Temporal Scales: USTEP employs a dual-gate mechanism that dynamically integrates information from both micro- and macro-temporal scales. USTEP achieves superior performance without incurring the high computational costs of recurrent-based models. For example: On Moving MNIST (Table~\ref{tab:equal}), USTEP w/TAU achieves an MSE of 21.84 with 17.7 GFLOPs, outperforming PredRNN++ (22.06 MSE with 171.7 GFLOPs) at only about 10\% additional computational cost compared to TAU. On KTH (Table~\ref{tab:kth40}), USTEP w/TAU achieves the best performance with an MSE of 54.68 at 214 GFLOPs, significantly outperforming PredRNN (MSE 61.42, 4730 GFLOPs) and PredRNN++ (MSE 74.21, 7032 GFLOPs). USTEP delivers this superior performance while using over 20$\times$ fewer FLOPs than recurrent-based models. 
\end{itemize}

USTEP bridges the gap between recurrent-based and recurrent-free methods by combining their strengths and addressing their limitations. It avoids the inefficiencies of frame-by-frame processing in recurrent-based models and overcomes the lack of fine-grained temporal modeling in recurrent-free models. This unified approach ensures that USTEP can handle both short-term and long-term dependencies dynamically and efficiently. As demonstrated in Section~\ref{sec:exp}, USTEP consistently achieves superior performance across benchmarks while maintaining a balanced computational cost, making it an effective and scalable solution for spatio-temporal predictive learning.

\subsection{Runtime Comparison on Different Devices}

To further emphasize the computational efficiency of our proposed framework, we conduct a detailed runtime comparison across different hardware platforms, using the KTH dataset with the extended frame prediction task ($10 \rightarrow 40$ frames) as a benchmark. This setup not only provides a challenging long-horizon forecasting scenario but also reveals the practical deployment costs across a range of devices, including high-performance GPUs (NVIDIA V100 32GB, NVIDIA A100 80GB) and a representative CPU platform (Intel(R) Xeon(R) Platinum 8358 @ 2.60GHz). Table~\ref{tab:kth_runtime} presents the training and inference times of USTEP, PredRNN++, and TAU on different platforms. On both GPU platforms, USTEP achieves nearly the same efficiency as TAU. Moreover, USTEP demonstrates significantly faster performance compared to PredRNN++.

\begin{table}[ht]
  \centering
  \caption{Runtime comparison on the KTH ($10 \rightarrow 40$ frames) dataset. Training time is measured in seconds per epoch, while inference time is measured in seconds of the test set.}
  \vspace{-2mm}
  \setlength{\tabcolsep}{3.2mm}{
  \begin{tabular}{ccccc}
  \toprule
  Device & Model & Training Time & Inference Time \\
  \midrule
  \multirow{3}{*}{V100} 
      & TAU        & 188   & 46   \\
      & PredRNN++  & 6360  & 1551 \\
      & USTEP      & 221   & 54   \\
  \hline
  \multirow{3}{*}{A100} 
      & TAU        & 101   & 25   \\
      & PredRNN++  & 3405  & 782  \\
      & USTEP      & 121   & 31   \\
  \hline
  \multirow{3}{*}{CPU} 
      & TAU        & 10528  & 2510  \\
      & PredRNN++  & 362520 & 88422 \\
      & USTEP      & 12155  & 2968  \\
  \bottomrule
  \end{tabular}}
  \label{tab:kth_runtime}
\end{table}

\section{Conclusion and Limitation}

This paper introduced USTEP, a novel paradigm for spatiotemporal prediction tasks, thoughtfully architected to unify the strengths of both recurrent-based and recurrent-free models. USTEP operates under a novel paradigm that offers a comprehensive view of spatiotemporal dynamics, facilitating a nuanced understanding and representation of intricate temporal patterns and dependencies. USTEP has proven its mettle across a variety of spatiotemporal tasks, demonstrating exceptional adaptability and superior performance in diverse contexts. It meticulously integrates local and global spatiotemporal information, providing a unified perspective that enhances its performance.

\begin{figure}[ht]
\centering
\includegraphics[width=0.48\textwidth]{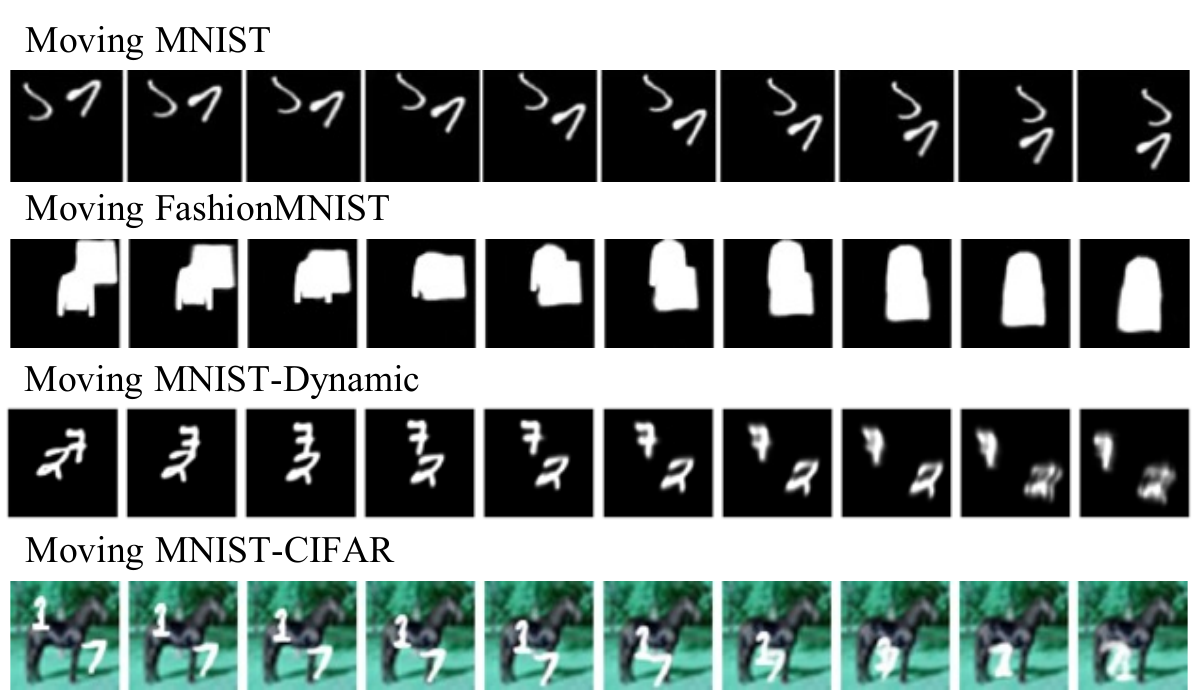}
\caption{Representative failure cases of USTEP on complex variants of Moving MNIST.}
\label{fig:failure}
\end{figure}

To better understand USTEP’s limitations, we conducted an in-depth failure analysis by designing three extended variants of the Moving MNIST that introduce controlled complexities: (i) Moving FashionMNIST: we replace digit characters with fashion items, increasing object variability and deformability; (ii) Moving MNIST-Dynamic: we introduce random Gaussian perturbations to the speed of each object, simulating non-uniform motion patterns; (iii) Moving MNIST-CIFAR: we use natural image backgrounds sampled from the CIFAR-10, introducing substantial background clutter and color variation. These variants are specifically constructed to test USTEP under conditions of occlusion, fast or irregular motion, and complex visual scenes. As illustrated in Fig.~\ref{fig:failure}, USTEP exhibits decreased predictions in these cases, particularly when object motion is highly erratic or when the foreground and background share similar textures or color distributions. In such settings, the model may produce blurred or temporally inconsistent predictions, revealing a sensitivity to dynamic and context-heavy environments. We believe these findings provide valuable insight into USTEP’s failure modes. In future work, we aim to address these challenges by incorporating adaptive motion modeling and stronger attention mechanisms for background-foreground separation.

\ifCLASSOPTIONcompsoc
  \section*{Acknowledgments}
\else
  \section*{Acknowledgment}
\fi

This work was supported by National Science and Technology Major Project (No. 2022ZD0115101), National Natural Science Foundation of China Project (No. 624B2115, No. U21A20427), Project (No. WU2022A009) from the Center of Synthetic Biology and Integrated Bioengineering of Westlake University, Project (No. WU2023C019) from the Westlake University Industries of the Future Research Funding.

\ifCLASSOPTIONcaptionsoff
  \newpage
\fi



%
\bibliographystyle{IEEEtran}
\bibliography{ref}
\end{document}